\documentclass{article}

 \usepackage[main, final]{neurips_2026}

\usepackage[utf8]{inputenc} 
\usepackage[T1]{fontenc}    
\usepackage{hyperref}       
\usepackage{url}            
\usepackage{booktabs}       
\usepackage{amsfonts}       
\usepackage{nicefrac}       
\usepackage{microtype}      
\usepackage{xcolor}         

\usepackage{graphicx}
\usepackage{amsmath}
\usepackage{amssymb}
\usepackage{mathtools}
\usepackage{amsthm}

\usepackage{wrapfig}
\usepackage{stfloats}
\usepackage{comment}
\usepackage{subcaption}
\usepackage{multirow}

\title{Shared Modular Recurrence in Contextual MDPs\\ for Universal Morphology Control}

\author{%
  Laurens Engwegen\\
  Delft University of Technology\\
  Delft, The Netherlands\\
  \texttt{l.r.engwegen@tudelft.nl}\\
  \And
  Max Weltevrede\\
  Delft University of Technology\\
  Delft, The Netherlands\\
  \texttt{m.r.weltevrede@tudelft.nl}\\
  \AND
  Caroline Horsch\\
  Delft University of Technology\\
  Delft, The Netherlands\\
  \texttt{c.c.h.horsch@tudelft.nl}\\
  \And
  Daan Brinks\\
  Delft University of Technology\\
  Delft, The Netherlands\\
  \texttt{d.brinks@tudelft.nl}\\
  \And
  Wendelin B\"ohmer\\
  Delft University of Technology\\
  Delft, The Netherlands\\
  \texttt{j.w.bohmer@tudelft.nl}\\
}

\begin{document}

\maketitle

\begin{abstract}
  A universal controller for any robot morphology would greatly improve computational and data efficiency. 
    Steps have been made towards such multi-robot control by utilizing \textit{contextual} information about the properties of individual robots and exploiting their modular structure in the architecture of deep reinforcement learning agents.
    When the robots have highly dissimilar morphologies, however, this becomes a challenging problem, especially when the agent must generalize to new, unseen robots.
    In this paper, we consider the fact that contextual features are often only partially available, but that they can be recovered through modular interactions.
    This can allow for better multi-robot control and generalization to contexts that are not seen during training. 
    To this extent, we implement a transformer-based architecture with shared modular recurrence and evaluate its (generalization) performance on a large set of MuJoCo robots. 
    The results show a substantial improvement in zero-shot generalization performance on robots with unseen dynamics, kinematics, and topologies, in four different environments.
\end{abstract}


\section{Introduction}

Reinforcement Learning (RL) has shown to be very promising for robotic control \citep{levine2016end,kalashnikov2018scalable,andrychowicz2020learning}. 
In an effort to close the gap with real-world applications, a lot of work has focused on RL agents that are able to generalize control to different tasks, e.g. manipulating different objects or acting in different environments.
Recently, large datasets of robot trajectories have been established and are being used to train such generalizable agents by learning from demonstrations in an offline fashion with foundation models \citep{RT1,RT2,openx}. 
Several works show promising possibilities for a single model to adapt not only to different scenes and goals, but also to different embodiments that the model has seen demonstrations of \citep{crossformer,octo}.
Zero-shot generalization to robots that were not seen during training, however, remains very challenging.

\begin{wrapfigure}{r}{0.55\textwidth}
    \centering
    \includegraphics[width=\linewidth]{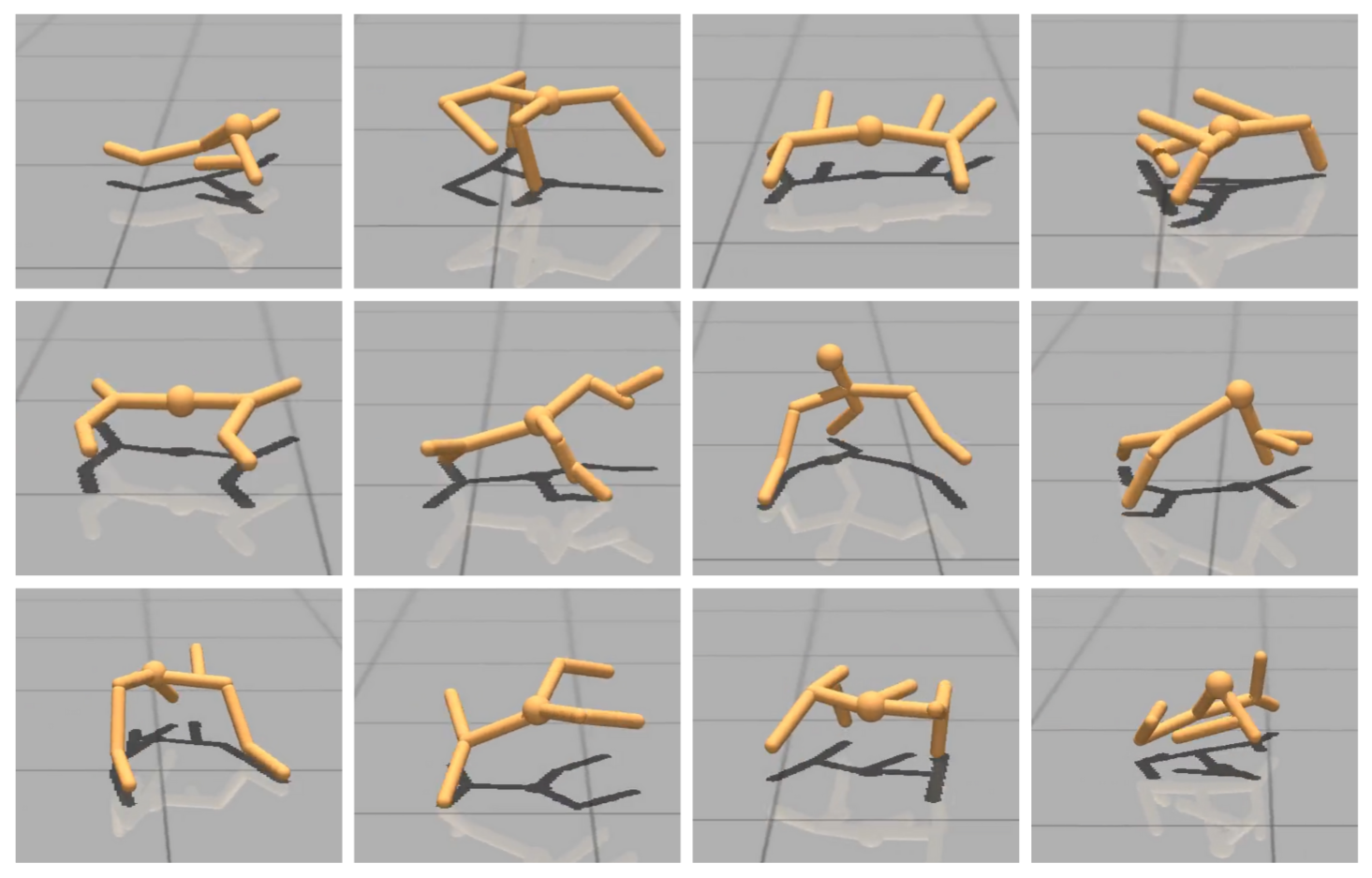}
  \caption{Example of robots that can be found in the UNIMAL design space \citep{unimal}.}
  \label{fig:robot_morphologies}
\end{wrapfigure}

Different robot embodiments, or \textit{morphologies}, can have advantages over each other, dependent on their task and the environments they act in.
It is infeasible to train a policy from scratch for every robot: training (or even fine-tuning) a policy for every new robot that we are interested in requires expensive compute, excessive use of data, and often tedious hyperparameter tuning.
A universal controller that can generalize control to any robot morphology could drastically improve efficiency. 
The UNIMAL design space \citep{unimal} was developed for the evolution of diverse robots that can perform varied tasks.
It contains more than 1000 different robots, a small subset of which is shown in Figure \ref{fig:robot_morphologies}.
The dataset of robots in the UNIMAL design space is suitable for the development and evaluation of RL agents that can generalize control to any robot.

By framing this problem as a multi-task RL problem \citep{MTRL_survey}, each robot can be considered as a separate task that has some specific \textit{contextual} features, such as the mass of its different limbs or its topology. 
The goal in this multi-task setting is to learn a universal controller that can control any robot on basis of its observations and context. 
This is not only challenging because robots can have different action and state spaces, but also because robots with different morphologies and/or dynamics might learn tasks in different ways. The multi-task framework allows us to evaluate the performance of an agent during multi-robot training, and its zero-shot generalization performance \citep{kirk_survey} to new, unseen robots.

Previous work on universal morphology control exploits contextual features that describe properties of the robot to better represent and distinguish tasks.
In this paper, we reason that the context can be considered partially observable and that the available features do not encapsulate the true state needed for optimal (generalizable) control.
We build upon modular architectures that were previously found to be effective for robotic control \citep{metamorph, modumorph}, 
and address the essentially unobservable context with shared modular recurrence to retain the system's compatibility with different action spaces while minimizing additional compute.
This paper empirically shows that the proposed system improves multi-robot control on varied training morphologies, and enables better zero-shot generalization to unseen test robots. 
Code to reproduce the experiments is publicly available at \url{https://github.com/LaurensEngwegen/smr}.

\section{Background}

\subsection{Contextual Markov Decision Process}

Here, the problem of learning RL policies that are trained on a set of training robots and must generalize to unseen test robots is considered. This problem can be formulated as a Markov Decision Process (MDP) where the agent can only partially access the MDP during training. An MDP is a tuple $( \mathcal{S}$, $\mathcal{A}$, $\mathcal{T}$, $\mathcal{R}$, $\rho )$, with state space $\mathcal{S}$, action space $\mathcal{A}$, transition function $\mathcal{T}(s_{t+1}|s_t,a_t)$, mapping current state $s_t \in \mathcal{S}$ and action $a_t \in \mathcal{A}$ to a probability distribution over next states $s_{t+1} \in \mathcal{S}$, reward function $\mathcal{R}: \mathcal{S} \times \mathcal{A} \rightarrow \mathbb{R}$ and initial state distribution $\rho (s_0)$.
A Contextual Markov Decision Process (CMDP \citep{hallak_cmdp}) is an MDP where states can be decomposed, $s_t = (s_t', c)$, into the underlying state $s_t' \in \mathcal{S'}$ and a context $c \in \mathcal{C}$.
The context is sampled at the start of each episode and remains static until the episode ends. In multi-robot control, the context defines the robot to be controlled. This framework allows to evaluate zero-shot generalization, that is, generalization to contexts not seen during training \citep{kirk_survey}.

The UNIMAL design space contains modular robots \citep{unimal} that are simulated with the MuJoCo physics engine \citep{mujoco}. Such robots consist of a set of nodes, or limbs, that share the same state space and action space, i.e. $\mathcal{S} = \{\mathcal{S}^i | i=1,...,N\}$ and $\mathcal{A} = \{\mathcal{A}^i | i=1,...,N\}$ for robots with $N$ limbs. 
Besides the underlying states, a node-level context, $\mathcal{C} = \{\mathcal{C}^i | i=1, ..., N\}$, is provided in the state that describes information of the limbs (e.g. mass, initial position with respect to the parent limb, and the initial position of each joint attached to the limb).
The node-level observation and context features that are provided to the agent are listed in Table \ref{tab:obs_context} in Appendix \ref{appendix:observation_context}.
Figure \ref{modular_robot_example} provides an example of a modular robot and how limb-wise observations, consisting of the underlying state $s'_{t}$ and (observable) context $c$, compose the global observation.

\begin{figure}
    \centering
    \includegraphics[width=0.6\linewidth]{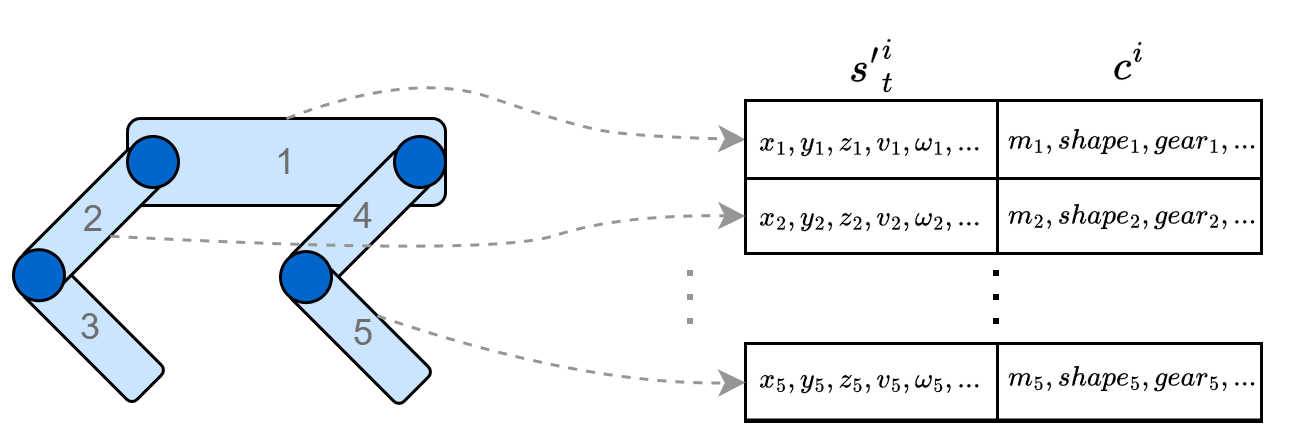}
    \caption{Example of how robot observations are modularized and can be decomposed into limb-wise underlying states and contexts.}
    \label{modular_robot_example}
\end{figure}

Previous methods exploited this modular structure in combination with modern architectures for effective multi-robot control \citep{SMP, amorpheus, metamorph, modumorph}. 
Rather than aiming to develop context-agnostic agents, \citet{metamorph} and \citet{modumorph} condition the agent on the available contextual information with methods coined \textit{MetaMorph} and \textit{ModuMorph}, respectively. In doing so, they did not only show improved performance during training, but also on generalization to unseen robots. 
However, the gap between the performance on training and testing robots remains substantial.

\subsection{Partially Observable Context}

In this paper, we recognize and leverage the fact that not all task-relevant contextual information is observable to the agent. In multi-agent control settings like MuJoCo, the missing context consists of structural and abstract properties that are difficult to encode directly in observations or cannot be processed effectively in the agent architecture.
First, structural information about the robot is not directly accessible. While the agent receives local positional relationships (e.g., a limb’s position relative to its parent), the true adjacency structure of the morphology is not explicitly provided. As noted by \citet{modumorph}, encoding such graph structure in modular architectures is not trivial due to challenges with positional encoding. Consequently, topology must be inferred indirectly. 
Moreover, some of the most critical contextual features are inherently abstract. For example, the functional influence that one limb exerts on another during coordinated movement is not directly observable nor easily encoded. 
Such effects of interaction cannot be captured through static observations alone.

Although the underlying state of a single robot in MuJoCo is treated as fully observable \citep{mujoco}, the CMDP is more accurately modeled as partially observable \citep{epistemicpomdp}. Formally, the emission or observation function defined in partially observable MDPs \citep{pomdp}, $\phi: \mathcal{S} \rightarrow \mathcal{O}$, maps a state to an observation $o_t \in \mathcal O$ that contains the underlying state and the observable context, $c^+$, for every timestep $t$: $o_t = \phi ((s_t', c)) = (s_t', c^+)$.
The unobserved portion of $c$ can include both missing physical parameters and abstract features.
In the single robot setting, such information never changes between episodes and can therefore effectively be treated as fully observable. Besides, even in the multi-robot setting, aforementioned methods that rely solely on explicitly provided context features can achieve good performance on training robots: the agent can simply overfit. The challenge arises when the agent must generalize to robots with (abstract) unobservable context features that were not present in the training set.

Incorporating memory into the agent architecture \citep{DRQN} can allow the agent to implicitly identify and quickly adapt to latent contextual information online. Because the underlying physical state is otherwise observable, any information inferred from history {\em must} be from the missing context.
Crucially, such memory mechanism must preserve the modular structure of the agent's architecture to remain compatible with the multi-robot setting in which input dimensionalities can vary. E.g., a naive encoding of the global action-observation history is not directly compatible with the state-of-the-art modular architectures.

\section{Related Work}

\subsection{Universal Morphology Control}
We aim to learn an RL policy that can generalize control to different robot morphologies, even when state and action dimensionalities can change.
Effective approaches utilized the modularity in robots \citep{DGN} and introduced weight sharing across different modules \citep{SMP}. 
Several works adopted graph neural networks \citep{GNN, nervenet} or transformers \citep{vaswani2017attention, amorpheus} as inductive biases to more explicitly infer relationships between different limbs or modules through message-passing.
More recently, multi-robot training has been evaluated on a larger scale of different morphologies with the introduction of the UNIMAL design space \citep{unimal}.
\citet{metamorph} constructed training and testing sets of robots to evaluate multi-robot control, and showed the effectiveness of a modular transformer-based approach when contextual information is provided to the agent.
By further exploiting this contextual information in the agent's architecture, \citet{modumorph} demonstrated improved performance.

Several recent works incorporate additional information about the robot topology through a graph encoding or sparse attention matrices \citep{getzero, bodytransformer} for improved generalization in simulation and the real world. 
Fine-tuning and policy distillation approaches have also been proposed for generalization purposes \citep{finetuning_metamorph, distilling}.
More complex modular architectures were introduced to improve transfer to unseen robots in simulation and the real world \citep{runthemall, MAT}, although evaluating on smaller sets of, and relatively similar robot morphologies. \citet{runthemall}, specifically, distinguish between different body parts (feet, torso, joints) for targeted processing. When this distinction cannot be made, like in the UNIMAL design space, the architecture is similar to MetaMorph with extra layers.
Our work builds upon MetaMorph and ModuMorph, while remaining compatible with the adaptations of the previously mentioned works that can be beneficial in other environments.

None of the aforementioned approaches consider this generalization problem as partially observable.
Earlier works did suggest that ``implicit system identification'' with memory-based policies can benefit robotic control, but did not utilize a modular system, nor evaluate generalization on varied morphologies \citep{recurrent_system_identification, recurrent_sim2real}.
LocoFormer \citet{locoformer} exploits long history contexts, spanning multiple episodes in a meta-learning fashion, to train generalist agents in an environment with procedurally generated robots. The Transformer-XL \citet{transformerxl} model in this setup requires significantly more computational resources, but the improved generalization serves as additional evidence that observation histories can be beneficial, also without a meta-learning setting.
We are the first to show that the addition of shared modular recurrence can improve (zero-shot generalization) performance on a diverse set of robot morphologies.

\subsection{Neural Architectures}
The effectiveness of the transformer architecture \citep{vaswani2017attention} for multi-robot control, lies in its capability to model pairwise dependencies between limbs with self-attention. 
Self-attention can be defined as $A = \sigma (QK^T/\sqrt{d}) V$, with query, key and value matrices $Q,K,V \in \mathbb{R}^{N \times d}$, for robots with $N$ limbs and a hidden size of $d$. Learnable parameters $W_Q, W_K$ and $W_V$ map the input $X \in \mathbb{R}^{N \times d}$ to those matrices, i.e. $Q = XW_Q$, $K = XW_K$ and $V = XW_V$, and $\sigma(\cdot)$ is a row-wise softmax function. 
In addition to the attention mechanism, \citet{modumorph} utilize hypernetworks \citep{ha2016hypernetworks} to more explicitly condition the agent on the available node-wise contextual information. Briefly, they train a hypernetwork that is conditioned on the observable context to generate (1) the parameters of a node-wise encoder that produces $X_V$ with which the value in the subsequent transformer encoder layer is calculated as $V = X_VW_V$, (2) $X_Q$ and $X_K$, where the query and key matrices are now defined as $Q = X_QW_Q$, $K = X_KW_K$, respectively, and (3) the parameters of a node-wise decoder that projects the output of the transformer encoder. 
Recurrent neural networks (RNNs), like LSTM and GRU \citep{LSTM, GRU}, are useful architectures to deal with partially observable domains in deep RL. \citet{sensoryneuron} combined LSTMs with attention to develop systems that can adapt to changes (permutations) of the input. 
Here, we build upon these methods to develop a transformer-based architecture with shared modular recurrence and hypernetworks to improve multi-robot control.

\section{Shared Modular Recurrence}\label{methods}

\subsection{Recurrent PPO}
The goal is to learn a universal control policy that is effective for any robot we can encounter in the UNIMAL design space by training on a limited set of $K$ robots. 
An effective approach to RL in partially observable domains is to learn an encoding of the belief over the agent's true state. 
This is often done by, at every timestep $t$, encoding the action-observation history (AOH) $\tau_t^k = (o_0^k, a_0^k, \dots, o_{t-1}^k, a_{t-1}^k, o_t^k)$ with an RNN, for (in the current domain) robot $k$ the agent is controlling.
In this way, the training objective can be formulated as finding parameters $\theta$ for policy $\pi_{\theta}(a_{t}^k | \tau_{t}^k )$ that maximize the (discounted) cumulative reward, averaged over all training robots: $\mathrm{max}_{\theta} \frac{1}{K} \sum_{k=1}^{K} \mathbb E_{\pi_\theta}[\sum_{t=0}^{H} \gamma^{t}r_{t}^{k}]$, with task horizon $H$. We implemented a recurrent version of Proximal Policy Optimization (PPO) \citep{ppo} to optimize this objective. 

To make recurrent training tractable at scale, we incorporate R2D2-style sequence-chunking and burn-in \citep{r2d2}, originally proposed for (off-policy) DRQN \citep{DRQN}, on the PPO rollout buffer.
Since episode lengths can vary from 1 timestep up to $H=1000$, storing complete, padded episodes for batch processing with RNNs is impractical and inefficient. Instead, we store overlapping chunks of episodes.
For each chunk of transitions, the RNN hidden states are initialized with the hidden states that are stored during rollout at the beginning of each chunk. After that, a burn-in phase is applied to reconstruct meaningful hidden states for the current parametrization. We use a chunk size of $m=80$ and a burn-in period of $l=20$, as those values were reported by \citet{r2d2} to be effective.
A default recurrent PPO setup for universal morphology control settings is insufficient: naively encoding global action-observation histories breaks both the compatibility with varying input dimensionalities and the structural invariances required for generalization. This motivates our architectural contribution.

\subsection{Shared Recurrent Network}

In order to retain the agent's modularity, we adopt and adapt a recurrent architecture that processes components of the input separately \citep{sensoryneuron}: every limb-level action and observation are processed individually through an RNN to encode local AOHs, $\tau_t^{i} = (o_0^{i}, a_0^{i}, \dots, o_{t-1}^{i}, a_{t-1}^{i}, o_t^{i})$, for every limb $i$ (we omit the superscript that indicates the robot as our policy is controlling only one robot at a time).
Since nodes share the same state space, the parameters of the RNN can be shared to increase scalability and retain modularity.
We only keep track of the hidden states of all limbs, $h_t^{i} = \mathrm{RNN}(o_t^i, a_{t-1}^i, h_{t-1}^i)$, that encode the local AOHs, $\tau^{i}$, which are initialized with zeros at the start of an episode.
In this way, the agent can approximate the relevant history, necessary for implicit inference of the true state, for each limb individually.
We build upon state-of-the art architectures for universal morphology control: MetaMorph \citep{metamorph} and ModuMorph \citep{modumorph}.

\begin{figure}[h]
    \centering
    \includegraphics[width=0.7\textwidth]{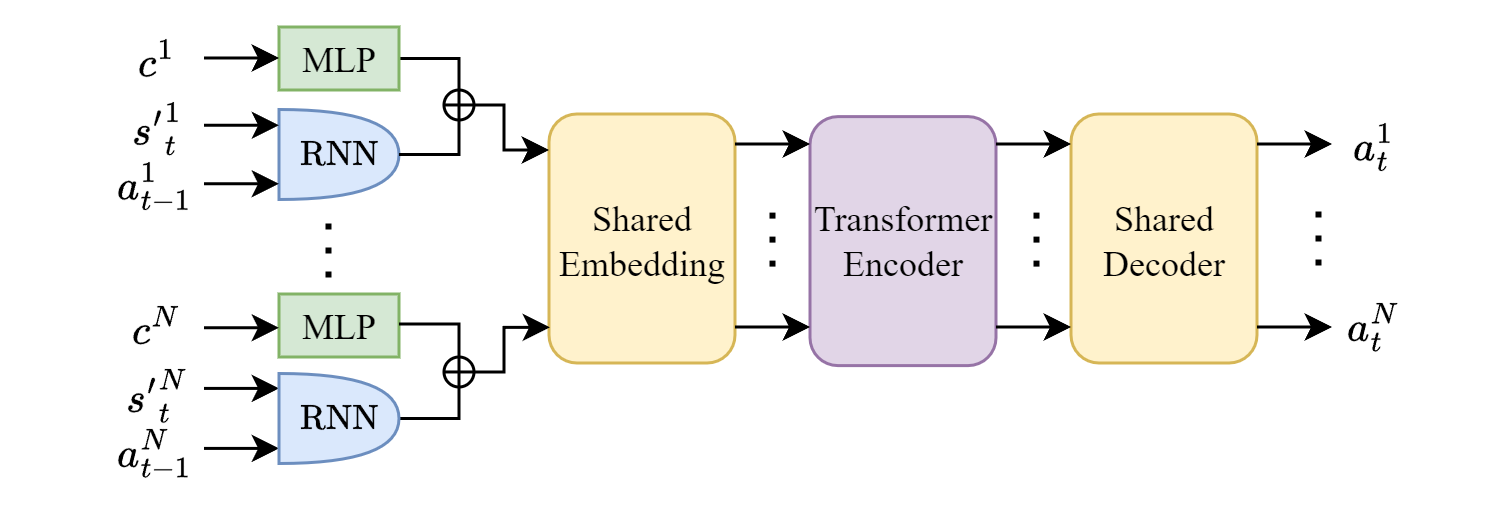}
    \caption{\textit{Recurrent MetaMorph (\textbf{R-MeMo})}}
    \label{fig:rmemo_architecture}
\end{figure}

\paragraph{Recurrent MetaMorph.} MetaMorph exploited a modular transformer-based architecture in which limb-wise observations are processed through a shared embedding layer, before applying multi-head attention. The resulting embeddings are mapped to actions with a shared decoder.
Consult Figure \ref{modular_robot_example} for a depiction of a limb-wise observation, decomposed into underlying state $s'^{i}_t$, and context $c^{i}$.
As shown in Figure \ref{fig:rmemo_architecture}, our recurrent version of MetaMorph (\textbf{R-MeMo}) first encodes the underlying state and previous action with an RNN. This is done separately for each limb, but with shared RNN parameters. The hidden state for each limb at the start of each episode chunk is stored in the rollout buffer for initialization during training. The observable context is mapped with an MLP, as this part of the observation remains constant throughout an episode. The embeddings are added together to establish a latent representation of $\tau_t$.

\begin{figure}[h]
    \centering
    \includegraphics[width=0.7\textwidth]{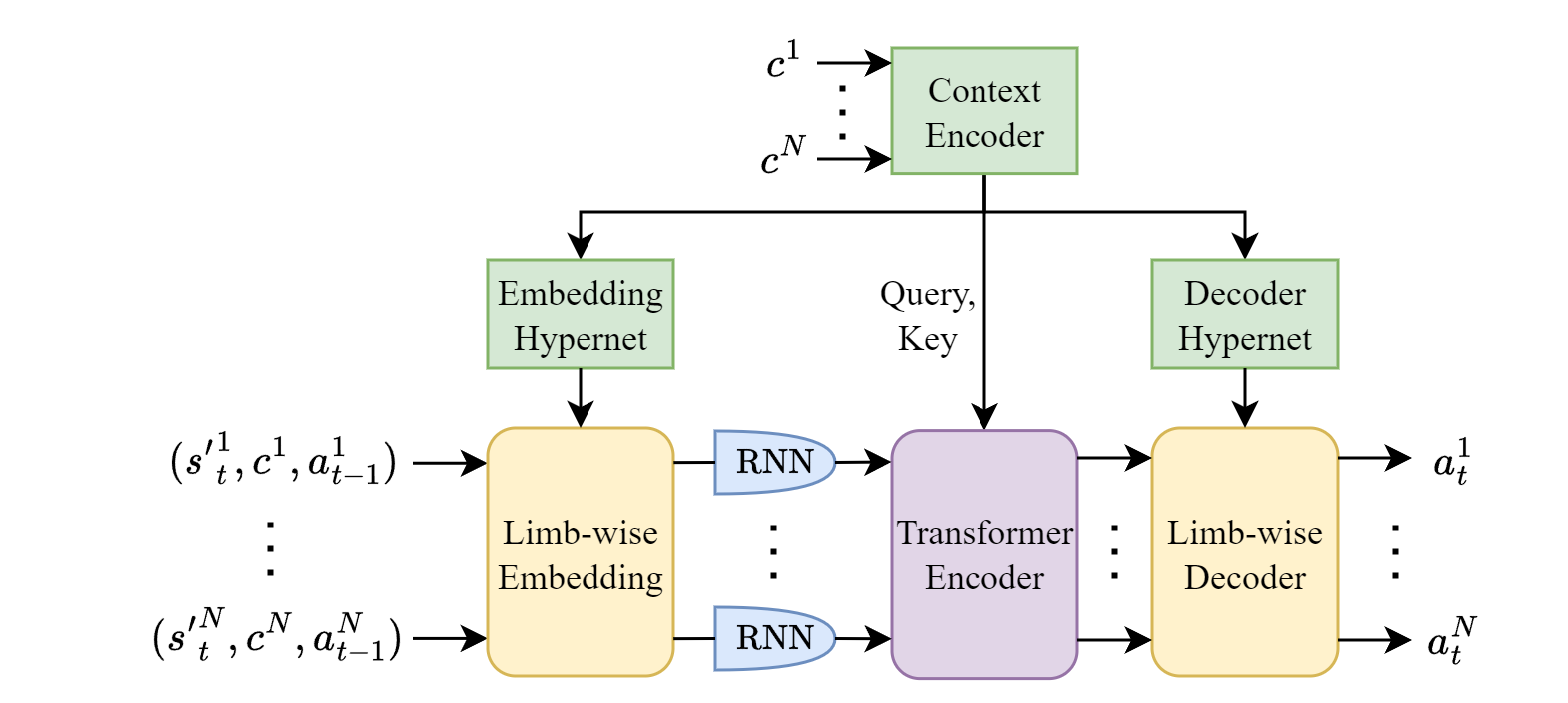}
    \caption{\textit{Recurrent ModuMorph (\textbf{R-MoMo})}}
    \label{fig:rmomo_architecture}
\end{figure}

\paragraph{Recurrent ModuMorph.} ModuMorph was built on MetaMorph by introducing a hypernetwork that takes context as input and generates the parameters for the embedding layer, the query and key of multi-head attention, and the parameters for the decoder layer.
Our recurrent ModuMorph version (\textbf{R-MoMo}), shown in Figure \ref{fig:rmomo_architecture}, uses the same hypernetwork and fixed attention as in the original architecture, but incorporates the shared modular RNN before the transformer to approximate local AOHs from latent encodings of the observation and previous action. 
The RNN could also be inserted before the embedding layer. This would, however, result in larger inputs for the embedding layer (if we keep the size of the hidden states the same): 128 instead of 54 per limb.
This, in turn, leads to an increase of more than 1M trainable parameters in the hypernetwork with respect to ModuMorph. This was avoided to ensure a fair comparison between ModuMorph and R-MoMo.

In both proposed architectures, a transformer receives the local (AOH) encodings from the RNN to infer relationships between different limbs and finally map them to limb-wise actions. The retained modular structure allows for robots with different input dimensionalities (i.e., different numbers of limbs). The introduction of shared modular recurrence only causes a modest increase in trainable parameters, as shown in Table \ref{tab:nr_parameters} in Appendix \ref{appendix:hyperparams}. Augmenting MetaMorph and ModuMorph with MLPs of similar size caused stability problems, demonstrated in Figure \ref{fig:train_extra_params} in Appendix \ref{appendix:hyperparams}, and was therefore avoided. Besides, the individual runs with additional parameters that did show stable learning did not achieve better performance.

\section{Experiments}\label{sec:experiments}

In this Section, experiments are performed with MetaMorph, ModuMorph, and their recurrent counterparts R-MeMo and R-MoMo, respectively. We use the MetaMorph version from \citet{modumorph}, which they report to perform better than the original implementation.

\subsection{Experimental setup}
Agents are trained with the dataset of 100 training robots as constructed by \citet{metamorph}.
We then first evaluate the agent's generalization performance on unseen variations of these training robots, where parameters that influence the dynamics and kinematics are altered (such as the damping of limbs or the angles joints can make). Subsequently, the performance on robots with unseen topologies is evaluated. 
The provided test set of robots with unseen topologies is randomly split into a validation (32 robots) and test (70 robots) set to experiment with different hyperparameters.

We only tuned one regularization hyperparameter that \citet{modumorph} found to have a big impact on performance. 
See Appendix~\ref{appendix:hyperparams} for further details. All other hyperparameter values were taken from \citet{modumorph} and are listed in Table \ref{tab:hyperparameters}. Parameters of the introduced shared modular recurrence were assigned default values (and, thus, not optimized), with a hidden and cell state size of 128 for the LSTM that was used as RNN. 
Figure \ref{fig:modumorph_fixedhistory} in Appendix \ref{appendix:ablations} shows that we cannot simply train ModuMorph with (fixed size) recent history as input to observe similar performance gains as with shared modular recurrence.
Other ablations, e.g., in which the LSTM is replaced by a GRU, and with 
recurrence in the hypernetwork, can also be found in Appendix \ref{appendix:ablations}.

Agents are trained and evaluated in four different environments. In each of those, the agent has to maximize the robot's locomotion distance. In \textbf{Flat Terrain}, the agent needs to traverse a flat surface, while in \textbf{Incline} the surface is inclined by 10 degrees. \textbf{Variable Terrain} contains a sequence of randomly generated hills, steps and rubble, interleaved with flat terrain. Finally, \textbf{Obstacles} is a flat terrain with randomly generated obstacles. 
In the latter two environments, the agent additionally receives a 2D heightmap of its close surrounding to be able to react to changes in terrain. 
For more details on the environments, we refer to \citet{unimal}.

\subsection{Multi-Robot Training Performance}

\begin{figure}[htbp]
    \centering
    \includegraphics[width=\textwidth]{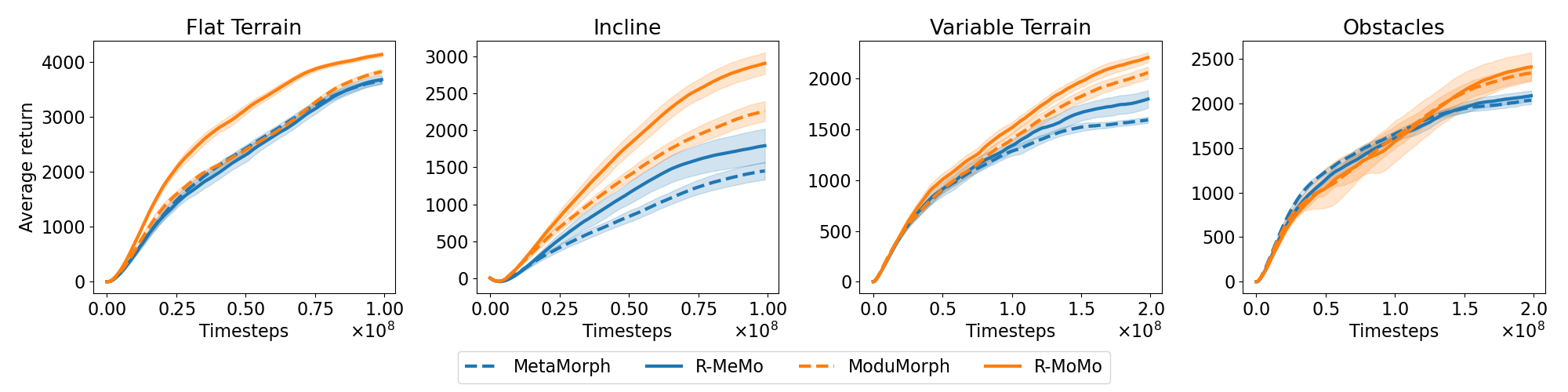}
    \caption{Training performance on the 100 training robots in the different environments. The average return with a 95\% confidence interval (CI) over 10 seeds is visualized.}
    \label{fig:train_performance}
\end{figure}

The training performance of the different methods on the 100 training robots, averaged over 10 seeds, is shown in Figure~\ref{fig:train_performance}. 
Across all environments, the recurrent architectures (R-MeMo and R-MoMo) obtain either similarly high or higher returns than their non-recurrent baselines MetaMorph and ModuMorph.
Particularly in the Incline environment, which is more difficult as dynamics play a more important role, shared modular recurrence results in better training performance.
In general, ModuMorph seems to perform better than MetaMorph and R-MeMo on the training robots, illustrating the effectiveness of the hypernetworks conditioned on the available context during multi-robot training. 
The training time of R-MeMo and R-MoMo increases slightly with respect to their baselines, which is reported in Appendix \ref{appendix:comparisons}. However, this is not the case during inference, as the agent then always processes observations sequentially.

\subsection{Zero-Shot Generalization to Different Dynamics and Kinematics}

\citet{metamorph} constructed a set of robots that have the same topologies as the training robots, but differ in a contextual feature, to evaluate zero-shot generalization to different dynamics or kinematics.
\begin{wrapfigure}{r}{0.6\textwidth}
    \centering
    \includegraphics[width=\linewidth]{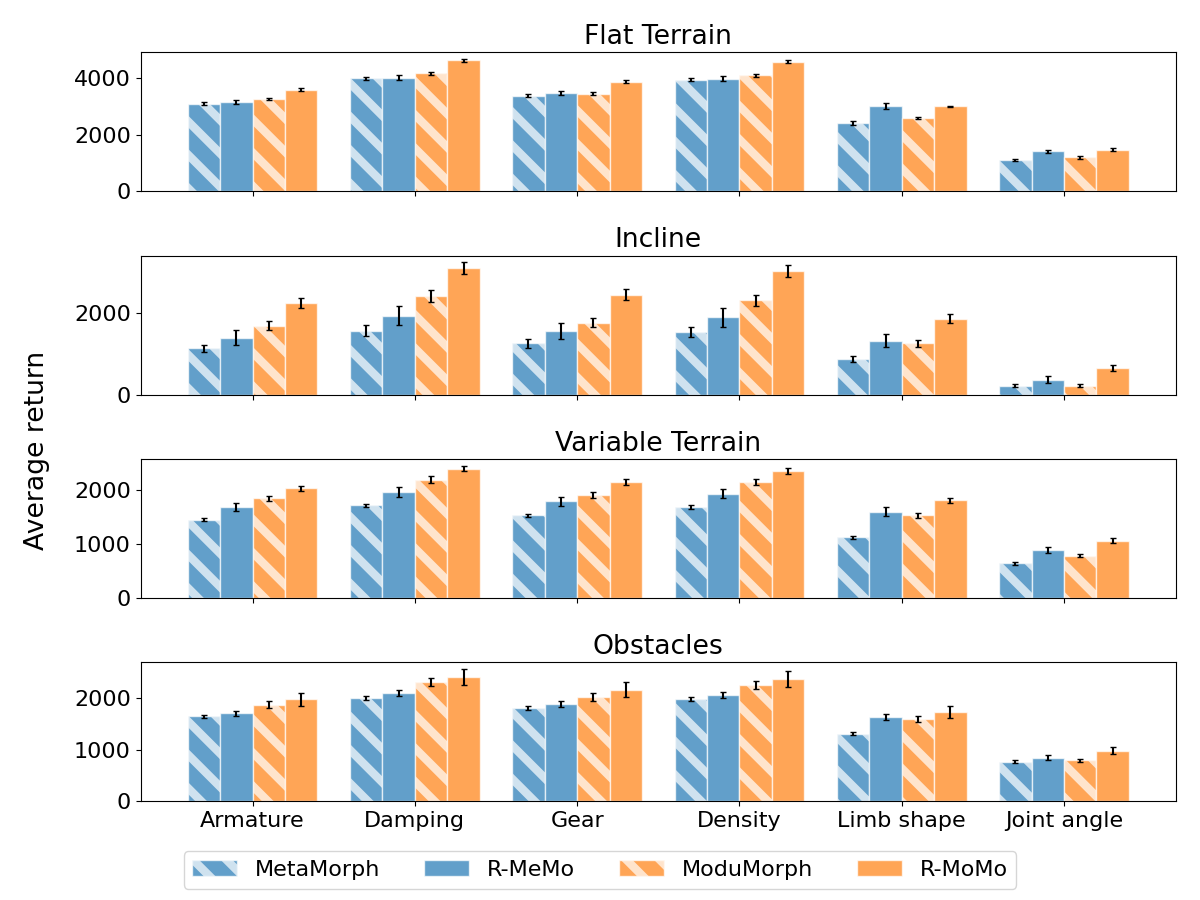}
    \caption{Performance on robots with changes in contextual features that result in unseen dynamics or kinematics. The average return with a 95\% CI over 10 seeds is reported.}
    \label{fig:dynamics_generalization}
    \vspace{-1ex}
\end{wrapfigure}
For each training robot, they created four test robots with variations in armature, damping, gear, density, limb shapes or joint angles, resulting in a new set of 2400 test robots. 
Figure \ref{fig:dynamics_generalization} shows the performance of the four evaluated methods on the test robots with unseen dynamics and kinematics parameters. Over all changes, R-MoMo obtains a higher average return than ModuMorph, consistently throughout the different environments. Only in Obstacles confidence intervals overlap. Similarly, R-MeMo performs better than MetaMorph, and in some environments and parameter changes even outperforms ModuMorph. These results demonstrate improved generalization performance obtained through the shared modular recurrence across environments with different levels of complexity.

\subsection{Zero-Shot Generalization to Unseen Robot Topologies}

\begin{figure}[htbp]
    \centering
    \includegraphics[width=0.8\linewidth]{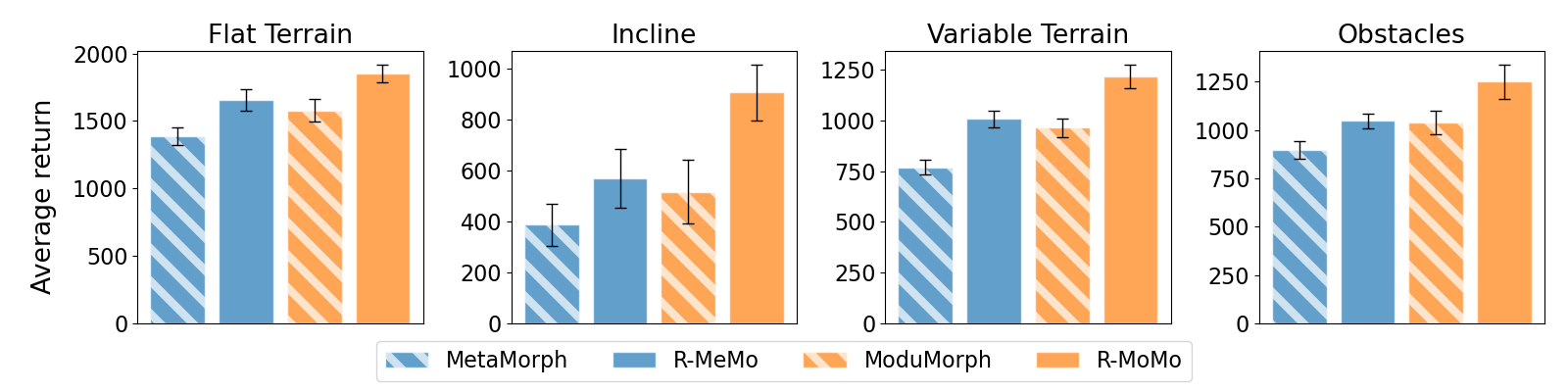}
    \caption{Test performance on robots with unseen topologies in four different environments. The average return with a 95\% CI is shown.}
    \label{fig:test_returns}
\end{figure}

Training and test robot topologies, $\mathcal{C}_{\mathrm{train}}$ and $\mathcal{C}_{\mathrm{test}}$, for which $\mathcal{C}_{\mathrm{train}} \cap \mathcal{C}_{\mathrm{test}} = \emptyset$, were constructed by \citet{metamorph}. These robots are very diverse, as they have a variable amount of limbs (up to 12), with different context features, and a variable amount of joints per limb (up to 2).
After training on the 100 training robots, the methods were evaluated on the 70 test robots with unseen topologies. 
The averaged returns of the different methods in the four environments are shown in Figure \ref{fig:test_returns}.
In each of the environments, ModuMorph and/or R-MoMo dominate training, but R-MoMo significantly outperforms the other methods in zero-shot generalization to the test robots. 
Additionally, R-MeMo is competitive with ModuMorph on the unseen test robots, even though ModuMorph performs better during training.
These results show that architectures with shared modular recurrence can learn policies that generalize much better to unseen test robots than their non-recurrent baselines across environments with varying levels of difficulty.

\begin{figure}[htbp]
    \centering    
    \begin{subfigure}[b]{0.7\linewidth}
        \includegraphics[width=\linewidth]{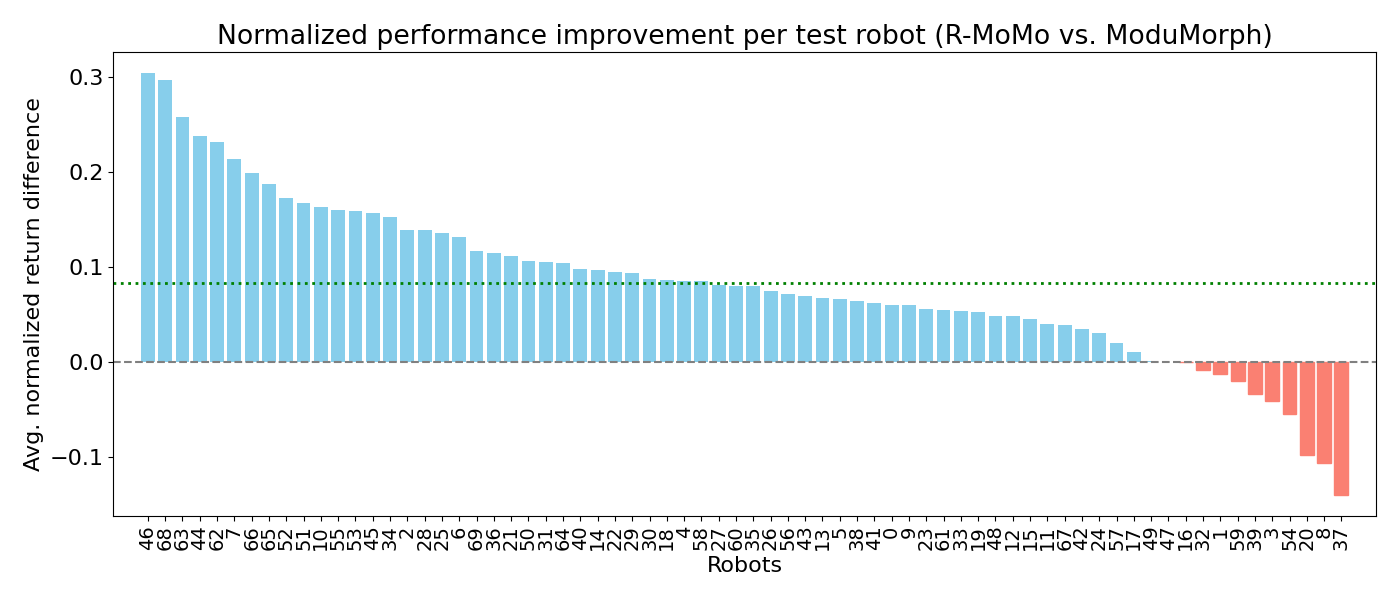}
    \end{subfigure}
    \begin{subfigure}[b]{0.7\linewidth}
        \includegraphics[width=\linewidth]{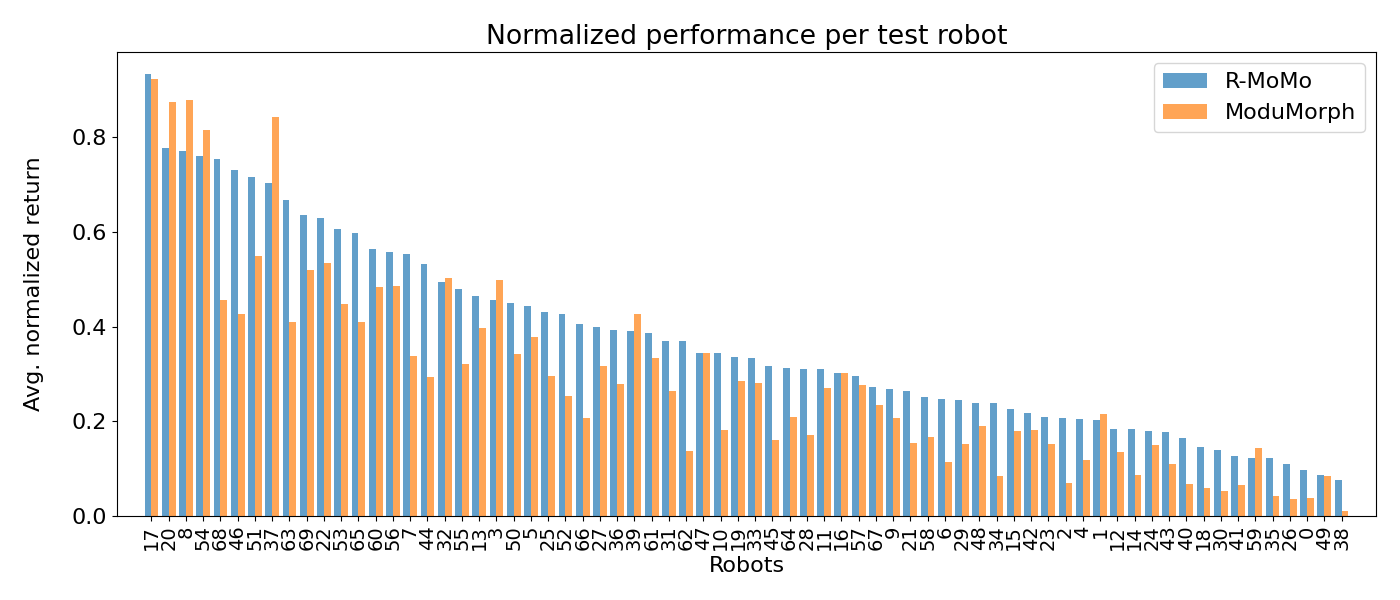}        
    \end{subfigure}
    \caption{The difference in return between R-MoMo and ModuMorph (top) and the obtained returns (bottom) on all test robots. The returns are normalized for each environment and averaged over 10 seeds. The green dotted line indicates the average performance improvement.}
    \label{fig:per_robot_test}
\end{figure}

Returns in each environment can range from very low ($<0$) to very high ($>4000$) values. 
Averaged returns over 70 test robots could therefore be misleading, as differences can be caused by only a small set of test robots. 
We therefore additionally report the difference in return between R-MoMo and ModuMorph for every test robot individually, next to the obtained returns, in Figure \ref{fig:per_robot_test}.
To compare the methods across all environments, the average returns are normalized using the minimal ($52$, $-74$, $71$, $126$) and maximal ($5008$, $3751$, $3000$, $2403$) returns found in Flat Terrain, Incline, Variable Terrain, and Obstacles, respectively. 
Per-robot performance comparisons for every environment separately can be found in Appendix \ref{appendix:comparisons}.
This comparison shows a consistent improvement in a majority of the test robots, rather than only on a limited set of robots.
Additionally, robots that ModuMorph performs better on are often also well controlled by R-MoMo. In contrast, ModuMorph struggles to control various robots across environments, while R-MoMo shows substantially fewer robots that are poorly controlled.
Analyses in Appendix \ref{appendix:comparisons} validate the consistency of the improved policy throughout the entire episode (rather than only at certain timesteps) for the agents with shared modular recurrence.

\subsection{Training Set Size}

\begin{wrapfigure}{r}{0.7\textwidth}
    \vspace{-2ex}
    \centering
    \includegraphics[width=\linewidth]{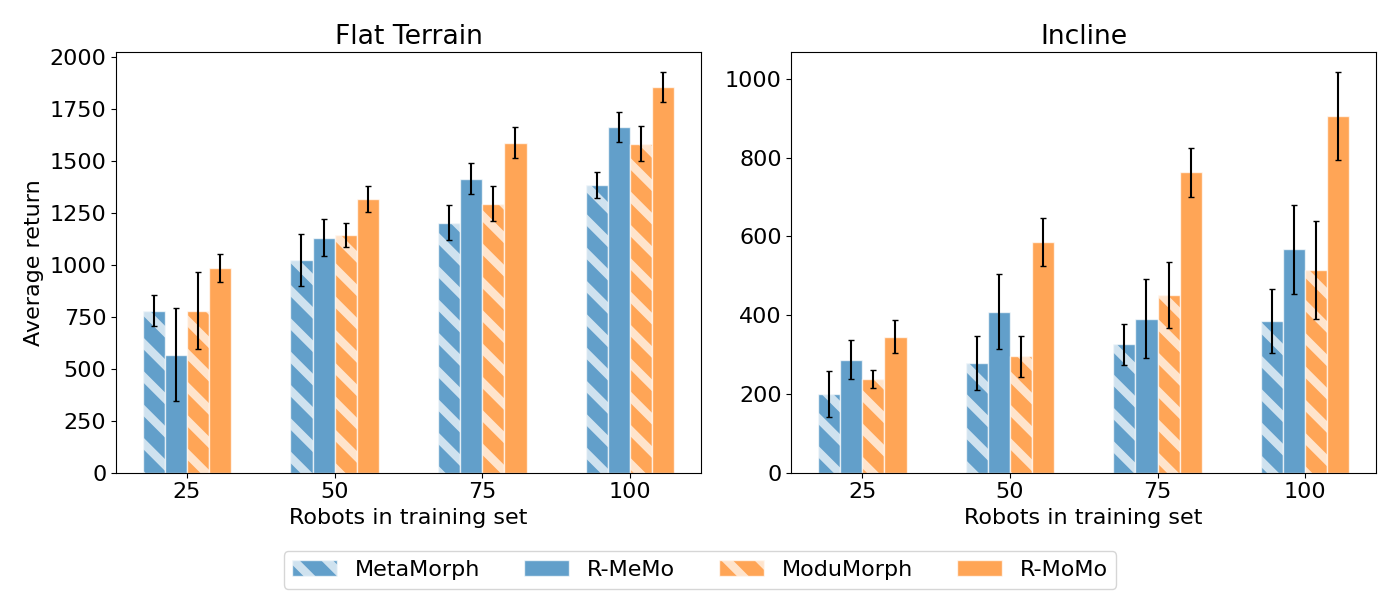}
    \caption{Test performance with smaller training set sizes, averaged over 10 seeds with 95\% CI error bars.}
    \label{fig:ablation_trainingsize}
\end{wrapfigure}
To further validate the increased generalization performance and scalability of the introduced approach, we performed experiments with different training set sizes.
Figure \ref{fig:ablation_trainingsize} shows the return obtained by the different methods for different amounts of robots in the training set: 25, 50, 75, and the complete set of 100 training robots in Flat Terrain and Incline. The improvement in zero-shot generalization performance of R-MoMo over ModuMorph is consistently visible for the different sizes of training sets, especially from 50 robots onwards. The difference increases when more training robots are available, which is specifically demonstrated in the more difficult Incline environment. This shows the consistency and scalability of R-MoMo with respect to the amount of training examples.
Furthermore, these results corroborate the scaling law hypothesis \citep{scalinglaws} stating that test performance scales with the number of available training embodiments. Interestingly, we do observe a difference in the slope of scaling across different methods: specifically in Incline, R-MoMo scales better than its baseline.

\subsection{Single Contextual Features}

\begin{figure}[htbp]
    \centering
    \begin{subfigure}[b]{0.45\linewidth}
        \includegraphics[width=\textwidth]{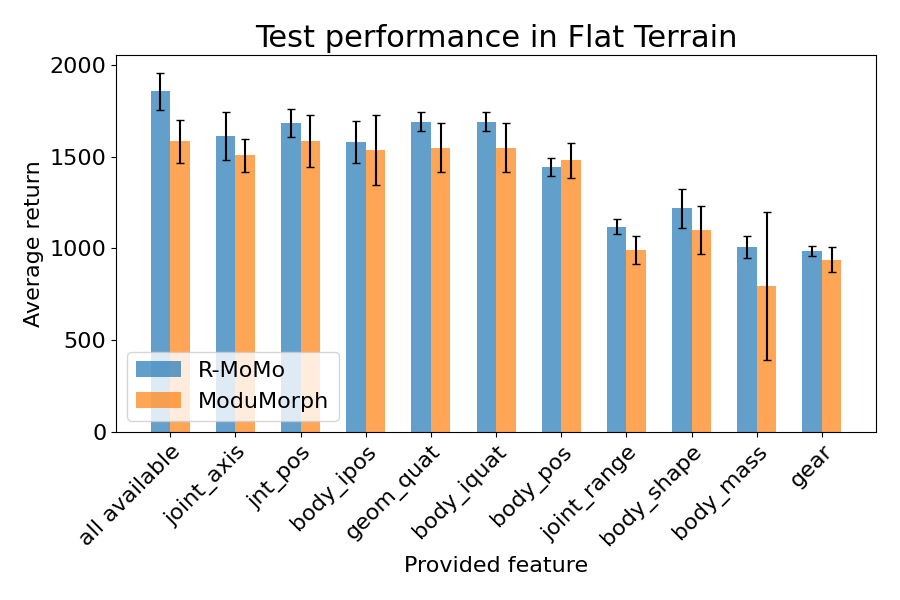}
    \end{subfigure}
    \begin{subfigure}[b]{0.45\linewidth}
        \includegraphics[width=\textwidth]{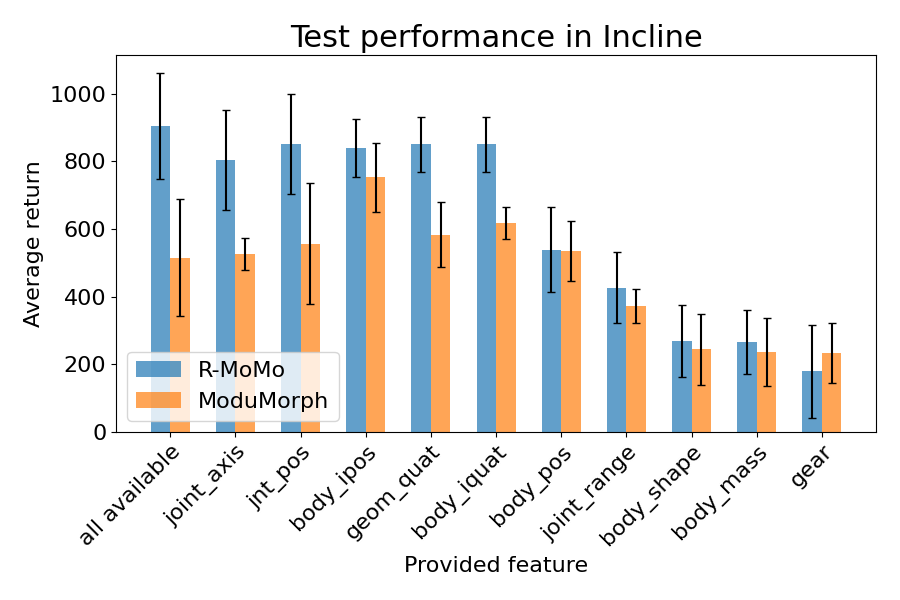}
    \end{subfigure}
    \caption{Test performance of ModuMorph and R-MoMo on Flat Terrain and Incline when providing either \textit{all available} contextual features or only a single contextual feature. Mean performance and standard deviation over 5 seeds is shown.}
    \label{fig:single_features}
\end{figure}

We experimented with a scenario in which a very limited number of contextual features would be available to the agent, to find potential differences in robustness against this lack of information. 
ModuMorph and R-MoMo were trained and evaluated under this constraint. Test performance using single contextual features (or using \textit{all available} features) in Flat Terrain and Incline is shown in Figure \ref{fig:single_features}.
Large error bars (e.g. for ModuMorph with $\mathrm{body\_mass}$ in Flat Terrain) correspond to highly instable performance across different random seeds.

These results show that R-MoMo frequently outperforms ModuMorph in the single feature setting. This observation is more evident in Incline, the more difficult environment.
However, this advantage is not consistent across all features: some features are insufficiently informative to induce strong performance when used in isolation for both methods.
When the provided contextual feature \textit{is} informative, R-MoMo's performance accelerates more than ModuMorph's. 
This sensitivity likely stems from the hypernetwork, which conditions exclusively on the available context. 
Notably, ModuMorph's performance in the sparse context regime does not drop for the first six features in both environments. A possible explanation is overfitting to the training robots, reducing the effect of the available feature.
Future work could investigate such sparse context availability and feature importance in more detail.

\section{Discussion}

This work explores the introduction of shared modular recurrence in transformer-based architectures for improved multi-robot control. 
It was hypothesized that modular recurrence could improve performance by enabling the agent to infer relevant latent context from local histories.
The results have shown a consistent increase in both training and zero-shot generalization performance when such memory mechanism was introduced across different environments for robots with different dynamics, kinematics, and topologies.
This clearly indicates that the RNN extracts some unobservable contextual information from the history for better policies.
Moreover, the improvement is consistent over a large set of test robots and for agents trained on varying amounts of training robots.

Experiments in which only a single context feature is provided did show an increase in performance, although not consistently for all features. 
Specifically in a more difficult environment, informative contextual features can enable the recurrent architecture to learn better policies. 
The results indicated that the implicit inference of more contextual information is dependent on the quality or informativeness of available context features. 
This is an interesting observation, as it implies that learning of partially observable contexts greatly benefits from observing some features, but not others. We leave it to future work to characterize what kind of features assist this inference best, and to what extent this is an artifact of the architecture's dependence on the hypernetwork.

A limitation of the explored recurrent architecture, is that hidden states have to be stored for each limb. 
Scaling to robots with a large number of limbs requires, therefore, more memory. 
Nonetheless, using episode chunking with a burn-in period during training only requires storing hidden states at the start of a chunk, minimizing the additional memory usage.
An interesting direction for future research would be to investigate a more efficient memory mechanism in the architecture.
Secondly, the sequential processing of RNNs results in longer training times.
This disadvantage only holds for training, due to the inherent sequential processing during inference.
Lastly, there is still room for improvement regarding the generalization gap.
Nonetheless, the combination of shared modular recurrence with transformers has shown to be promising for multi-robot control and could be effective in other problems with partially observable contexts in graph-like structures.


\begin{ack}
This project was funded by the Delft AI initiative within the BIOlab. This work was also partially funded by the Dutch Research Council (NWO) project {\em Reliable Out-of-Distribution Generalization in Deep Reinforcement Learning} with project number OCENW.M.21.234. DB acknowledges support from an ERC Starting Grant (850818 - MULTI-VIsion) and the Convergence Health and Technology (Integrative Neuromedicine flagship). Computational resources in this work were provided by the \citet{DAIC} at TU Delft (RRID: SCR\_025091).
\end{ack}


\bibliographystyle{plainnat}
\bibliography{bibliography}

@misc{DAIC,
  author = {{Delft AI Cluster (DAIC)}},
  title = {{The Delft AI Cluster (DAIC), RRID:SCR\_025091}},
  year = {2024},
  doi = {10.4233/rrid:scr_025091},
  url = {https://daic.tudelft.nl/}
}

@article{MTRL_survey,
  title={A survey of multi-task deep reinforcement learning},
  author={Vithayathil Varghese, Nelson and Mahmoud, Qusay H},
  journal={Electronics},
  volume={9},
  number={9},
  pages={1363},
  year={2020},
  publisher={MDPI}
}

@incollection{pomdp,
  title={Partially observable Markov decision processes},
  author={Spaan, Matthijs TJ},
  booktitle={Reinforcement learning: State-of-the-art},
  pages={387--414},
  year={2012},
  publisher={Springer}
}

@inproceedings{DRQN,
  title={Deep Recurrent {Q}-Learning for Partially Observable {MDP}s},
  author={Hausknecht, Matthew J and Stone, Peter},
  booktitle={2015 AAAI Fall Symposia},
  volume={45},
  pages={141},
  year={2015}
}

@article{LSTM,
  title={Long short-term memory},
  author={Hochreiter, Sepp and Schmidhuber, J{\"u}rgen},
  journal={Neural Computation},
  volume={9},
  number={8},
  pages={1735--1780},
  year={1997},
  publisher={MIT press}
}

@inproceedings{GRU,
  title={On the properties of neural machine translation: Encoder--decoder approaches},
  author={Cho, Kyunghyun and Van Merri{\"e}nboer, Bart and Bahdanau, Dzmitry and Bengio, Yoshua},
  booktitle={Proceedings of SSST-8, eighth workshop on syntax, semantics and structure in statistical translation},
  pages={103--111},
  year={2014}
}

@article{levine2016end,
  title={End-to-end training of deep visuomotor policies},
  author={Levine, Sergey and Finn, Chelsea and Darrell, Trevor and Abbeel, Pieter},
  journal={Journal of Machine Learning Research},
  volume={17},
  number={39},
  pages={1--40},
  year={2016}
}

@inproceedings{kalashnikov2018scalable,
  title={Scalable deep reinforcement learning for vision-based robotic manipulation},
  author={Kalashnikov, Dmitry and Irpan, Alex and Pastor, Peter and Ibarz, Julian and Herzog, Alexander and Jang, Eric and Quillen, Deirdre and Holly, Ethan and Kalakrishnan, Mrinal and Vanhoucke, Vincent and others},
  booktitle={Conference on robot learning},
  pages={651--673},
  year={2018},
  organization={PMLR}
}

@article{andrychowicz2020learning,
  title={Learning dexterous in-hand manipulation},
  author={Andrychowicz, OpenAI: Marcin and Baker, Bowen and Chociej, Maciek and Jozefowicz, Rafal and McGrew, Bob and Pachocki, Jakub and Petron, Arthur and Plappert, Matthias and Powell, Glenn and Ray, Alex and others},
  journal={The International Journal of Robotics Research},
  volume={39},
  number={1},
  pages={3--20},
  year={2020},
  publisher={SAGE Publications Sage UK: London, England}
}

@article{RT1,
  title={{RT}-1: Robotics transformer for real-world control at scale},
  author={Brohan, Anthony and Brown, Noah and Carbajal, Justice and Chebotar, Yevgen and Dabis, Joseph and Finn, Chelsea and Gopalakrishnan, Keerthana and Hausman, Karol and Herzog, Alex and Hsu, Jasmine and others},
  journal={arXiv preprint arXiv:2212.06817},
  year={2022}
}

@inproceedings{RT2,
  title={{RT}-2: Vision-language-action models transfer web knowledge to robotic control},
  author={Zitkovich, Brianna and Yu, Tianhe and Xu, Sichun and Xu, Peng and Xiao, Ted and Xia, Fei and Wu, Jialin and Wohlhart, Paul and Welker, Stefan and Wahid, Ayzaan and others},
  booktitle={Conference on Robot Learning},
  pages={2165--2183},
  year={2023},
  organization={PMLR}
}

@inproceedings{openx,
  title={Open {X}-embodiment: Robotic learning datasets and {RT-X} models: Open {X}-embodiment collaboration$^0$},
  author={O’Neill, Abby and Rehman, Abdul and Maddukuri, Abhiram and Gupta, Abhishek and Padalkar, Abhishek and Lee, Abraham and Pooley, Acorn and Gupta, Agrim and Mandlekar, Ajay and Jain, Ajinkya and others},
  booktitle={2024 IEEE International Conference on Robotics and Automation (ICRA)},
  pages={6892--6903},
  year={2024},
  organization={IEEE}
}

@article{crossformer,
  title={Scaling cross-embodied learning: One policy for manipulation, navigation, locomotion and aviation},
  author={Doshi, Ria and Walke, Homer and Mees, Oier and Dasari, Sudeep and Levine, Sergey},
  journal={arXiv preprint arXiv:2408.11812},
  year={2024}
}

@inproceedings{octo,
    title={Octo: An Open-Source Generalist Robot Policy},
    author = {{Octo Model Team} and Dibya Ghosh and Homer Walke and Karl Pertsch and Kevin Black and Oier Mees and Sudeep Dasari and Joey Hejna and Charles Xu and Jianlan Luo and Tobias Kreiman and {You Liang} Tan and Lawrence Yunliang Chen and Pannag Sanketi and Quan Vuong and Ted Xiao and Dorsa Sadigh and Chelsea Finn and Sergey Levine},
    booktitle = {Proceedings of Robotics: Science and Systems},
    address  = {Delft, Netherlands},
    year = {2024},
}

@article{recurrent_system_identification,
  title={Preparing for the unknown: Learning a universal policy with online system identification},
  author={Yu, Wenhao and Tan, Jie and Liu, C Karen and Turk, Greg},
  journal={arXiv preprint arXiv:1702.02453},
  year={2017}
}

@inproceedings{recurrent_sim2real,
  title={Sim-to-real transfer of robotic control with dynamics randomization},
  author={Peng, Xue Bin and Andrychowicz, Marcin and Zaremba, Wojciech and Abbeel, Pieter},
  booktitle={2018 IEEE international conference on robotics and automation (ICRA)},
  pages={3803--3810},
  year={2018},
  organization={IEEE}
}

@inproceedings{locoformer,
  title={LocoFormer: Generalist locomotion via long-context adaptation},
  author={Liu, Min and Pathak, Deepak and Agarwal, Ananye},
  booktitle={Conference on Robot Learning},
  pages={532--546},
  year={2025},
  organization={PMLR}
}

@inproceedings{transformerxl,
  title={Transformer-xl: Attentive language models beyond a fixed-length context},
  author={Dai, Zihang and Yang, Zhilin and Yang, Yiming and Carbonell, Jaime G and Le, Quoc and Salakhutdinov, Ruslan},
  booktitle={Proceedings of the 57th annual meeting of the association for computational linguistics},
  pages={2978--2988},
  year={2019}
}

@inproceedings{scalinglaws,
  title={Towards embodiment scaling laws in robot locomotion},
  author={Ai, Bo and Dai, Liu and Bohlinger, Nico and Li, Dichen and Mu, Tongzhou and Wu, Zhanxin and Fay, K and Christensen, Henrik I and Peters, Jan and Su, Hao},
  booktitle={Conference on Robot Learning},
  pages={3483-3515},
  year={2025},
  organization={PMLR}
}

@article{getzero,
  title={Get-zero: Graph embodiment transformer for zero-shot embodiment generalization},
  author={Patel, Austin and Song, Shuran},
  journal={arXiv preprint arXiv:2407.15002},
  year={2024}
}

@article{bodytransformer,
  title={Body transformer: Leveraging robot embodiment for policy learning},
  author={Sferrazza, Carmelo and Huang, Dun-Ming and Liu, Fangchen and Lee, Jongmin and Abbeel, Pieter},
  journal={arXiv preprint arXiv:2408.06316},
  year={2024}
}

@article{finetuning_metamorph,
  title={Efficient Morphology-Aware Policy Transfer to New Embodiments},
  author={Przystupa, Michael and Tang, Hongyao and Jagersand, Martin and Miret, Santiago and Phielipp, Mariano and Taylor, Matthew E and Berseth, Glen},
  journal={Reinforcement Learning Journal},
  year={2025}
}

@article{distilling,
  title={Distilling morphology-conditioned hypernetworks for efficient universal morphology control},
  author={Xiong, Zheng and Vuorio, Risto and Beck, Jacob and Zimmer, Matthieu and Shao, Kun and Whiteson, Shimon},
  journal={International Conference on Machine Learning},
  year={2024}
}

@article{runthemall,
  title={One policy to run them all: an end-to-end learning approach to multi-embodiment locomotion},
  author={Bohlinger, Nico and Czechmanowski, Grzegorz and Krupka, Maciej and Kicki, Piotr and Walas, Krzysztof and Peters, Jan and Tateo, Davide},
  journal={arXiv preprint arXiv:2409.06366},
  year={2024}
}

@article{MAT,
  title={Mat: Morphological adaptive transformer for universal morphology policy learning},
  author={Li, Boyu and Li, Haoran and Zhu, Yuanheng and Zhao, Dongbin},
  journal={IEEE Transactions on Cognitive and Developmental Systems},
  volume={16},
  number={4},
  pages={1611--1621},
  year={2024},
  publisher={IEEE}
}

@article{hallak_cmdp,
      title={Contextual Markov Decision Processes}, 
      author={Assaf Hallak and Dotan Di Castro and Shie Mannor},
      year={2015},
      journal={arXiv preprint arXiv:1502.02259}
}

@article{kirk_survey,
  title={A survey of zero-shot generalisation in deep reinforcement learning},
  author={Kirk, Robert and Zhang, Amy and Grefenstette, Edward and Rockt{\"a}schel, Tim},
  journal={Journal of Artificial Intelligence Research},
  volume={76},
  pages={201--264},
  year={2023}
}

@article{GNN,
  title={The graph neural network model},
  author={Scarselli, Franco and Gori, Marco and Tsoi, Ah Chung and Hagenbuchner, Markus and Monfardini, Gabriele},
  journal={IEEE transactions on neural networks},
  volume={20},
  number={1},
  pages={61--80},
  year={2008},
  publisher={IEEE}
}

@article{vaswani2017attention,
  title={Attention is all you need},
  author={Vaswani, Ashish and Shazeer, Noam and Parmar, Niki and Uszkoreit, Jakob and Jones, Llion and Gomez, Aidan N and Kaiser, {\L}ukasz and Polosukhin, Illia},
  journal={Advances in Neural Information Processing Systems},
  volume={30},
  year={2017}
}

@article{ha2016hypernetworks,
  title={Hypernetworks},
  author={Ha, David and Dai, Andrew and Le, Quoc V},
  journal={International Conference on Learning Representations},
  year={2017}
}

@article{sensoryneuron,
  title={The sensory neuron as a transformer: Permutation-invariant neural networks for reinforcement learning},
  author={Tang, Yujin and Ha, David},
  journal={Advances in Neural Information Processing Systems},
  volume={34},
  pages={22574--22587},
  year={2021}
}

@inproceedings{SMP,
  title={One policy to control them all: Shared modular policies for agent-agnostic control},
  author={Huang, Wenlong and Mordatch, Igor and Pathak, Deepak},
  booktitle={International Conference on Machine Learning},
  pages={4455--4464},
  year={2020},
  organization={PMLR}
}

@article{DGN,
  title={Learning to control self-assembling morphologies: a study of generalization via modularity},
  author={Pathak, Deepak and Lu, Christopher and Darrell, Trevor and Isola, Phillip and Efros, Alexei A},
  journal={Advances in Neural Information Processing Systems},
  volume={32},
  year={2019}
}

@inproceedings{nervenet,
  title={Nervenet: Learning structured policy with graph neural networks},
  author={Wang, Tingwu and Liao, Renjie and Ba, Jimmy and Fidler, Sanja},
  booktitle={International Conference on Learning Representations},
  year={2018}
}

@article{amorpheus,
  author       = {Vitaly Kurin and
                  Maximilian Igl and
                  Tim Rockt{\"{a}}schel and
                  Wendelin B{\"{o}}hmer and
                  Shimon Whiteson},
  title        = {My Body is a Cage: the Role of Morphology in Graph-Based Incompatible
                  Control},
  journal      = {International Conference on Learning Representations},
  year         = {2021},
  NOurl          = {https://arxiv.org/abs/2010.01856},
  eprinttype    = {arXiv},
  eprint       = {2010.01856},
  bibsource    = {dblp computer science bibliography, https://dblp.org}
}

@article{metamorph,
  title={Metamorph: Learning universal controllers with transformers},
  author={Gupta, Agrim and Fan, Linxi and Ganguli, Surya and Fei-Fei, Li},
  journal={International Conference on Learning Representations},
  year={2022}
}

@inproceedings{modumorph,
  title={Universal morphology control via contextual modulation},
  author={Xiong, Zheng and Beck, Jacob and Whiteson, Shimon},
  booktitle={International Conference on Machine Learning},
  pages={38286--38300},
  year={2023},
  organization={PMLR}
}

@article{epistemicpomdp,
  title={Why generalization in {RL} is difficult: Epistemic {POMDP}s and implicit partial observability},
  author={Ghosh, Dibya and Rahme, Jad and Kumar, Aviral and Zhang, Amy and Adams, Ryan P and Levine, Sergey},
  journal={Advances in Neural Information Processing Systems},
  volume={34},
  pages={25502--25515},
  year={2021}
}

@inproceedings{mujoco,
  title={{MuJoCo}: A physics engine for model-based control},
  author={Todorov, Emanuel and Erez, Tom and Tassa, Yuval},
  booktitle={2012 IEEE/RSJ international conference on intelligent robots and systems},
  pages={5026--5033},
  year={2012},
  organization={IEEE}
}

@article{unimal,
  title={Embodied intelligence via learning and evolution},
  author={Gupta, Agrim and Savarese, Silvio and Ganguli, Surya and Fei-Fei, Li},
  journal={Nature Communications},
  volume={12},
  number={1},
  pages={5721},
  year={2021},
  publisher={Nature Publishing Group UK London}
}

@article{ppo,
  title={Proximal policy optimization algorithms},
  author={Schulman, John and Wolski, Filip and Dhariwal, Prafulla and Radford, Alec and Klimov, Oleg},
  journal={arXiv preprint arXiv:1707.06347},
  year={2017}
}

@inproceedings{r2d2,
  title={Recurrent experience replay in distributed reinforcement learning},
  author={Kapturowski, Steven and Ostrovski, Georg and Quan, John and Munos, Remi and Dabney, Will},
  booktitle={International Conference on Learning Representations},
  year={2018}
}


\appendix

\newpage

\section{Observations and Context}\label{appendix:observation_context}

The observation provided to the agent in the current environments consists of various features for each limb. Table \ref{tab:obs_context} lists all of these features with a short description, the dimensionality of the feature and whether the feature is part of the context. We refer to the MuJoCo documentation \citep{mujoco} for more details.

\begin{table}[htbp]
    \centering
    \caption{Features of observations and context with description and dimensionality. $*$ indicates that each limb contains this feature twice, as every limb can contain two joints.}
    \begin{tabular}{llcc}
        
        \toprule
        \textbf{Feature} & \textbf{Description} & \textbf{Dim.} & \textbf{Context} \\
        
        \midrule
         $ \mathrm{body\_xpos} $ & Current x,y,z position of each limb & 3 & No \\
         $ \mathrm{body\_xvelp} $ & Linear velocity of each limb & 3 & No \\
         $ \mathrm{body\_xvelr} $ & Angular velocity of each limb & 3 & No \\
         $ \mathrm{body\_xquat} $ & Orientation of each limb & 4 & No \\
         $ \mathrm{qpos} $ & Generalized coordinates of each joint & 1* & No \\
         $ \mathrm{qvel} $ & Generalized velocity of each joint & 1* & No \\ 

         \midrule
         $ \mathrm{body\_pos} $ & Initial x,y,z position of limb w.r.t. parent limb & 3 & Yes \\
         $ \mathrm{body\_ipos} $ & Initial x,y,z position of center of mass w.r.t. parent limb & 3 & Yes \\
         $ \mathrm{body\_iquat} $ & Inverse quaternion of limb orientation & 4 & Yes \\
         $ \mathrm{geom\_quat} $ & Quaternion of geom relative to the body & 4 & Yes \\
         $ \mathrm{body\_mass} $ & Limb mass & 1 & Yes \\
         $ \mathrm{body\_shape} $ & Limb shape & 2 & Yes \\
         $ \mathrm{jnt\_pos} $ & Initial (x,y,z) coordinate of each joint & 3* & Yes \\
         $ \mathrm{joint\_range} $ & Range of motion (lower and upper bound) of each joint & 2* & Yes \\
         $ \mathrm{joint\_axis} $ & Axis of rotation/translation of each joint (one-hot for x,y,z)& 3* & Yes \\
         $ \mathrm{gear} $ & Gear ratio for each joint & 1* & Yes\\
         \bottomrule
    \end{tabular}
    \label{tab:obs_context}
\end{table}

\begin{figure}[htbp]
    \centering
    \begin{subfigure}[b]{0.7\textwidth}
        \includegraphics[width=\textwidth]{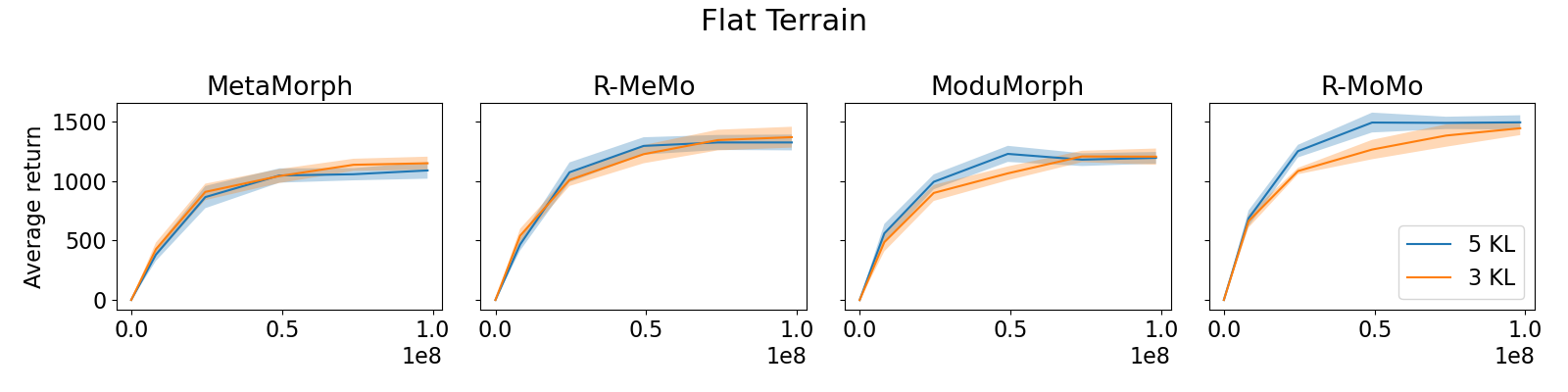}
    \end{subfigure}
    \begin{subfigure}[b]{0.7\textwidth}
        \includegraphics[width=\textwidth]{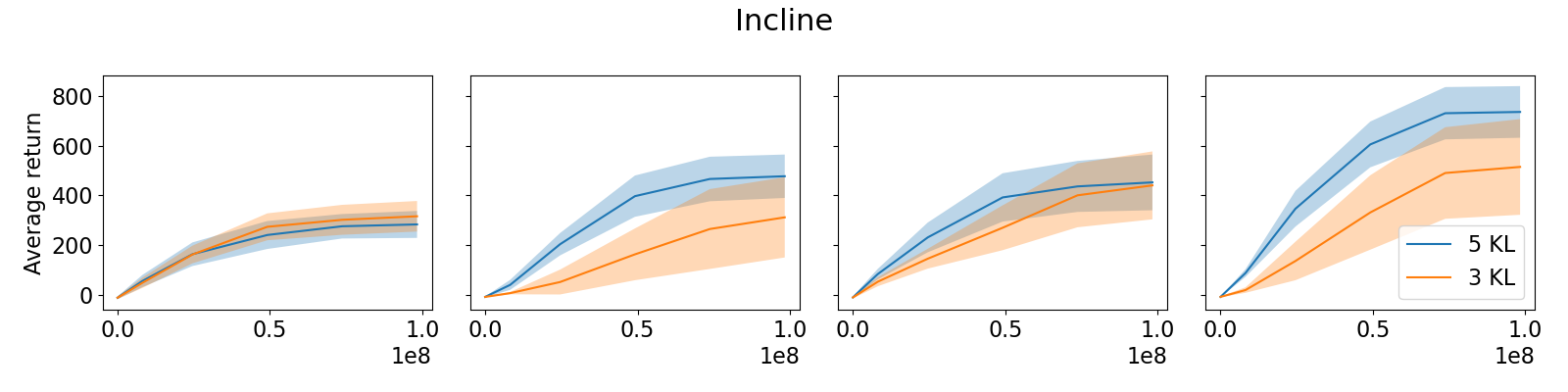}
    \end{subfigure}
    \begin{subfigure}[b]{0.7\textwidth}
        \includegraphics[width=\textwidth]{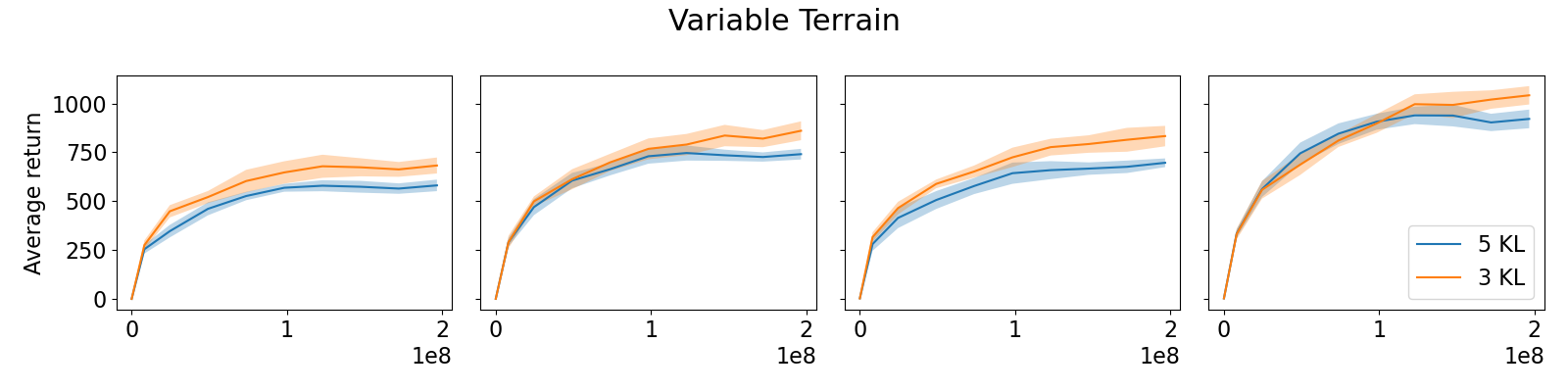}
    \end{subfigure}
    \begin{subfigure}[b]{0.7\textwidth}
        \includegraphics[width=\textwidth]{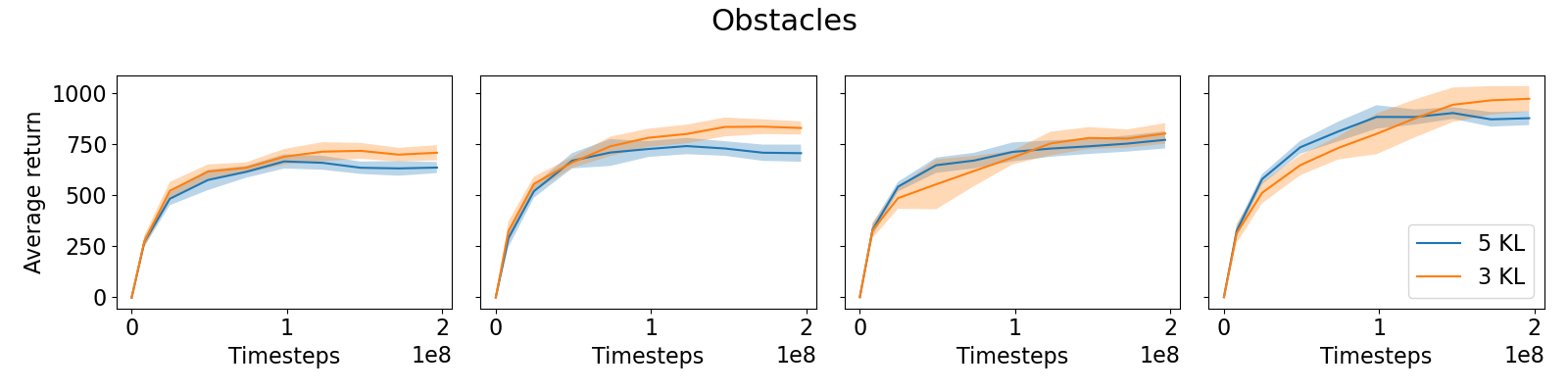}
    \end{subfigure}
    \caption{Validation performance on the 32 robots in the validation set for different values of the regularization parameter that defines the maximum approximate KL-divergence between the old and the new policy, averaged over 10 seeds with 95\% CIs.}
    \label{fig:validationset}
\end{figure}

\section{Implementation Details}\label{appendix:hyperparams}

\subsection{Regularization}

In our experiments, we use the same hyperparameter values as in ModuMorph (see Table \ref{tab:hyperparameters}). We only evaluate two different values for a regularization parameter on the validation set. \citet{modumorph} namely argued that this parameter can have a big impact on performance. This parameter defines the maximum approximate KL-divergence between the old and the new policy for every mini-batch before the update step. If this value is exceeded, the iteration of updates ends, and we return to sampling new trajectories. Figure \ref{fig:validationset} shows the average performance of the different methods with the two validated values ($3 \times 10^{-2}$ and $5 \times 10^{-2}$) that were also evaluated in ModuMorph. In most cases, there is not a big difference in performance. For every method, the value that resulted in the highest average return on the validation set was used for the reported results in Section \ref{sec:experiments}.

\subsection{Hyperparameters}

\citet{modumorph} provide a link to the datasets of robots and to their implementation on which we built. Table \ref{tab:hyperparameters} lists all the hyperparameter values that are used for all four methods that are considered in this paper. These values correspond to those used by \citet{modumorph}.
At the start of each episode rollout, a robot is sampled following the \textit{dynamic replay buffer balancing scheme} introduced by \citet{metamorph} (and used in both MetaMorph and ModuMorph). Note that the RNN is only used in R-MeMo and R-MoMo.

\begin{table}[htbp]
    \centering
    \caption{Hyperparameters used in the experiments with all methods.}
    \begin{tabular}{clr}
    
        \toprule
        & \textbf{Hyperparameter} & \textbf{Value} \\
        
        \midrule
        \multirow{2}{*}{Environment}
        & Total timesteps &  $\{1\times10^8, 2\times10^8\}$ \\
        & Vectorized environments & 32 \\

        \midrule
        \multirow{12}{*}{PPO}
        & Discount factor ($\gamma$) & 0.99 \\
        & GAE parameter ($\lambda$) & 0.95 \\
        & PPO clip range ($\epsilon$) & 0.2  \\
        & Reward normalization & Yes \\
        & Observation normalization & Yes \\
        & Batch size & 5120 \\
        & Timesteps per rollout & 2560 \\
        & Updates per rollout & 8 \\        
        & Gradient clipping & 0.5 \\
        & Clipped value function & Yes \\
        & Value loss coefficient & 0.5 \\
        
        \midrule
        \multirow{5}{*}{Optimizer}
        & Type & Adam \\
        & Epsilon & $1 \times 10^{-5}$ \\
        & Initial learning rate & $3 \times 10^{-4}$ \\
        & Learning rate schedule & Linear warmup, cosine decay \\
        & Warmup iterations & 5 \\

        \midrule
        \multirow{6}{*}{Transformer}
        & Number of layers & 5 \\
        & Number of attention heads & 2 \\
        & Embedding dimension & 128 \\
        & Feedforward dimension & 1024 \\
        & Activation function & ReLU \\
        & Dropout & 0.0 \\

        \midrule
        \multirow{4}{*}{RNN}
        & Type & LSTM \\
        & Number of layers & 1 \\
        & Embedding dimension & 128 \\
        & Dropout & 0.0 \\
        \bottomrule
        
    \end{tabular}
    \label{tab:hyperparameters}
\end{table}

\subsection{Increased Trainable Parameters}

\begin{figure}[htbp]
    \centering
    \begin{subfigure}[htbp]{0.45\linewidth}
        \includegraphics[width=\linewidth]{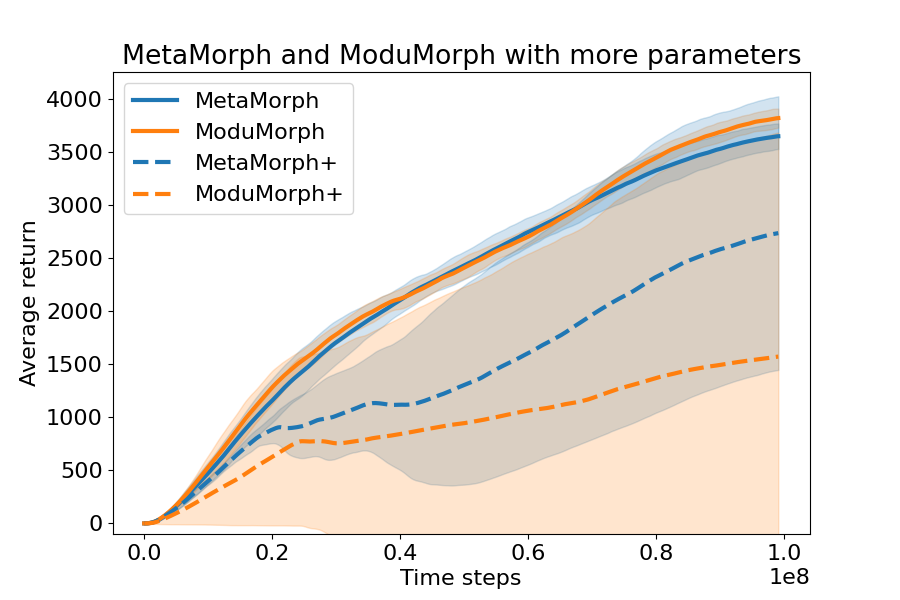}
    \end{subfigure}
    \begin{subfigure}[htbp]{0.45\linewidth}
        \includegraphics[width=\linewidth]{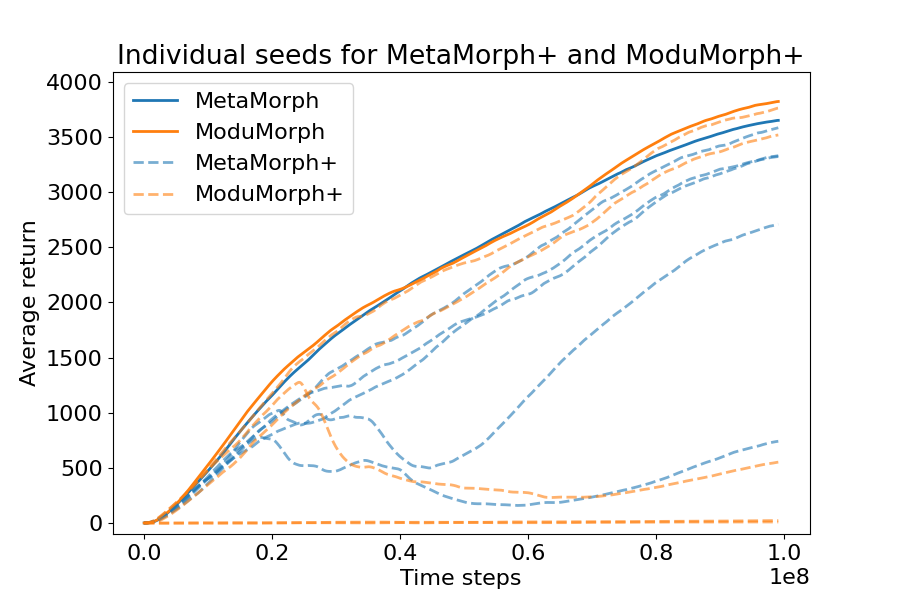}
    \end{subfigure}
    \caption{Training performance of MetaMorph and ModuMorph with additional trainable parameters (MetaMorph+ and ModuMorph+), comparable to the amount of additional parameters in the recurrent architectures, in Flat Terrain. The average return over 5 seeds with 95\% CIs is shown on the left. On the right, individual seeds are shown for MetaMorph+ and ModuMorph+, which demonstrates that none of the stable runs perform better than the average of the baselines.}
    \label{fig:train_extra_params}
\end{figure}

Table \ref{tab:nr_parameters} shows the number of trainable parameters for all of the methods. The shared modular recurrence introduces a small increase. We experimented with versions of MetaMorph and ModuMorph with a similar number of additional trainable parameters. For MetaMorph, the first embedding block is increased by two fully connected layers (with a hidden size of 256 and ReLU activation). In ModuMorph these layers were added after the first embedding block, where the RNN is placed in R-MoMo. Figure \ref{fig:train_extra_params} shows that the extra trainable parameters cause instability issues during training. Therefore, we stick to the architectures with the original number of parameters for the other experiments reported in this paper.

\begin{table}[htbp]
    \centering
    \caption{Number of trainable parameters for each method.}
    \begin{tabular}{lc}
        
        \toprule
         & \textbf{Trainable parameters} \\
        
        \midrule
        MetaMorph & 3,312,515 \\
        R-MeMo & 3,610,499 \\
        ModuMorph & 5,232,774 \\
        R-MoMo & 5,563,526 \\
        \bottomrule
    \end{tabular}
    \label{tab:nr_parameters}
\end{table}

\section{Ablations}\label{appendix:ablations}

\subsection{ModuMorph with Fixed History Window}

\begin{figure}[h]
    \centering
    \includegraphics[width=0.6\textwidth]{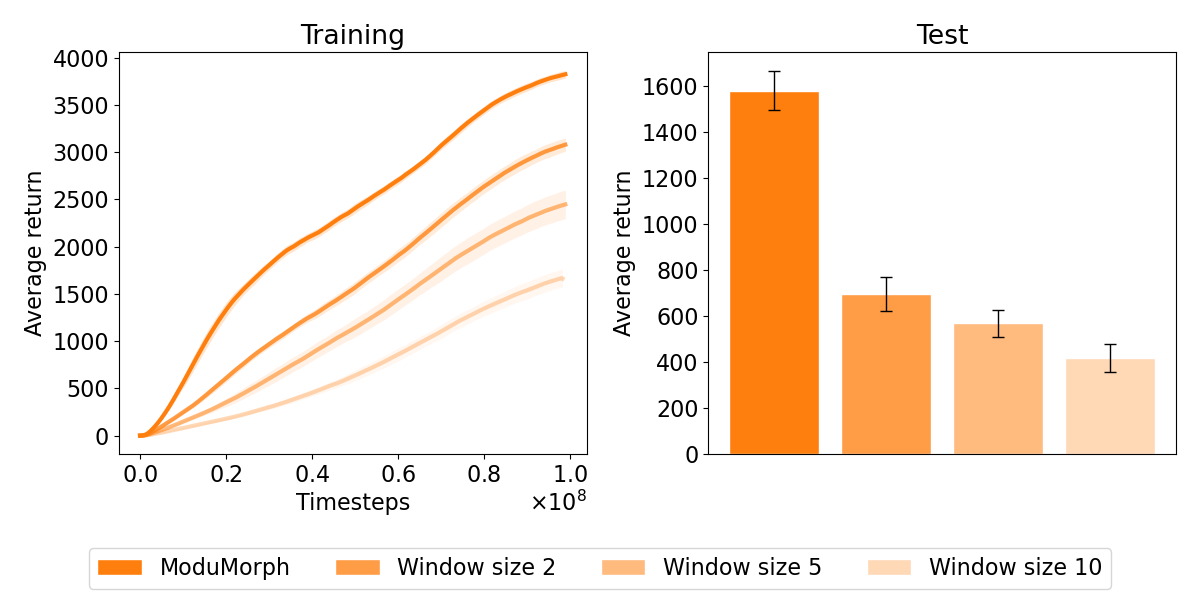}
    \caption{Training (left) and test (right) performance of ModuMorph with different window sizes of past observations (original 1, 2, 5, or 10) in Flat Terrain. Average return with a 95\% CI over 10 seeds is shown.}
    \label{fig:modumorph_fixedhistory}
\end{figure}

Here we show the performance of versions of ModuMorph that receive a fixed (small) window of previous observations and actions to investigate to what extent the RNN is necessary to capture (long range, generalizable) temporal dependencies or whether a recurrent architecture is not required to find the observed performance differences.
Note that stacking multiple observations significantly increases the number of trainable parameters, especially in ModuMorph. Instead of 6,784 parameters to encode a single limb's observation, we need 13,952 parameters for two observations and previous actions. This, in turn, leads to 1,785,856 trainable parameters in the hypernetwork instead of 868,352 (that map a 128-size context embedding to encoder parameters). This quickly explodes for larger windows of observations. \\
Figure \ref{fig:modumorph_fixedhistory} shows performance during training on the left and final test performance on the 70 unseen robot topologies on the right in the Flat Terrain environment. We experiment with default ModuMorph (i.e., window size 1), a history window size of 2, 5, and 10. The results show that performance degrades with larger history windows.
It is plausible that this is due to the combination of the larger and more varied input spaces and the substantial increase in parameters, which can require the network to train longer before converging to a good solution, and could make the network more susceptible to overfitting.

\begin{wrapfigure}{r}{0.55\textwidth}
    \centering
    \includegraphics[width=0.55\textwidth]{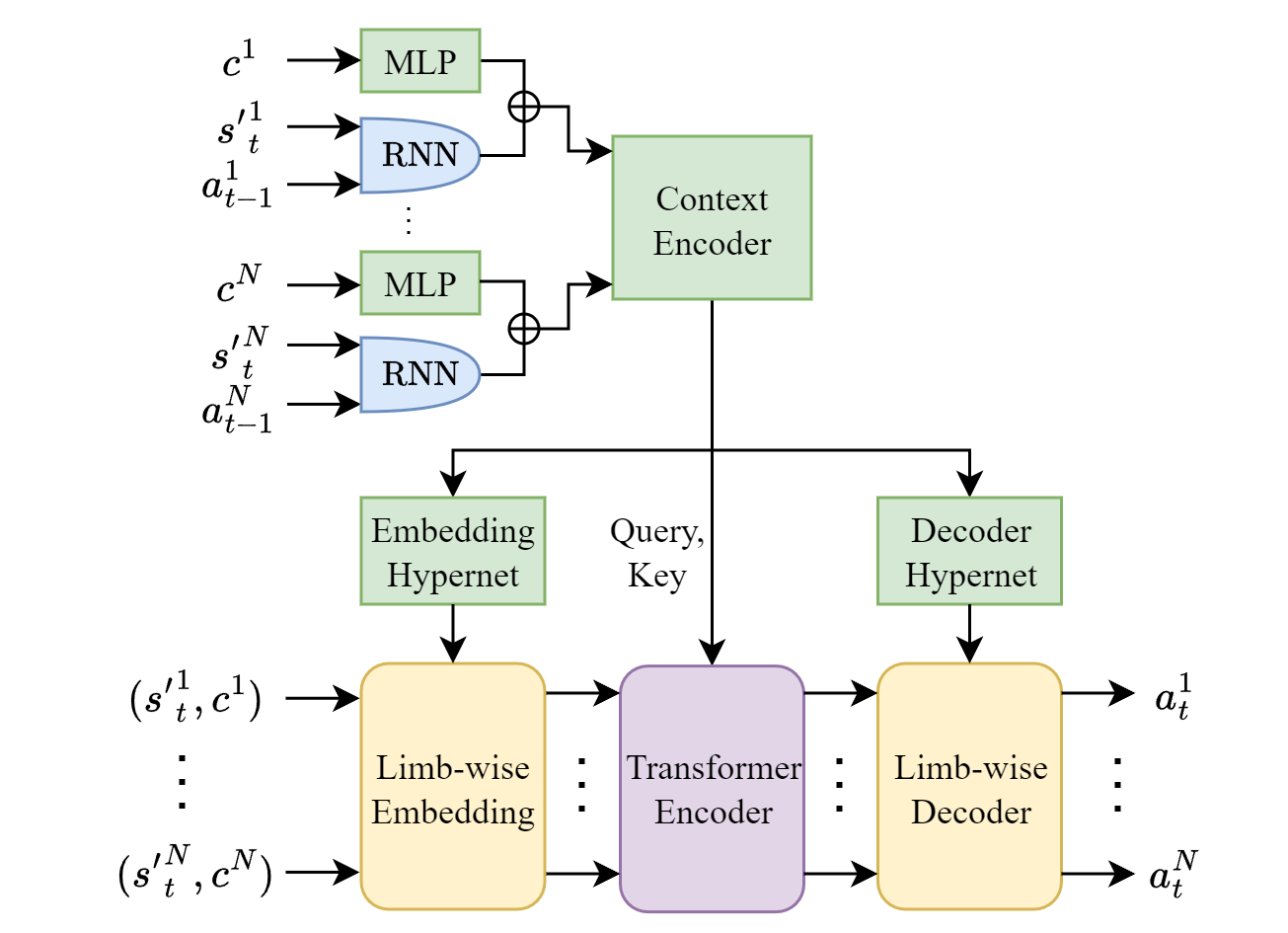}
    \caption{The HyperR-MoMo architecture variation, in which the AOH representation is learned in the hypernetwork instead of in the base network.}
    \label{fig:hyperrmomo_architecture}
\end{wrapfigure}

\subsection{Modular Recurrence in Hypernetwork}
Instead of inserting the RNN (here: LSTM) in the base controller, it can also be introduced in the hypernetwork to learn to infer latent contextual information. In this case, we add the AOH representation to the mapped observable context, to form the input for the blocks that generate the embedding layer, the query and key, and the decoder layer. The original RNN in the base controller (i.e. before the Transformer) is removed. 
We refer to this variation as HyperR-MoMo and illustrate it in Figure \ref{fig:hyperrmomo_architecture}.
Note that the original hypernetwork only required a forward pass at the start of an episode, since the context does not change during an episode. HyperR-MoMo's hypernetwork requires a forward pass at every timestep and can, thus, generate different parameters at every timestep.

The training and zero-shot generalization performance of this architecture variation in Flat Terrain and Incline is shown in Figures \ref{fig:ablation_training_performance} and \ref{fig:ablation_test_performance}, respectively. In Flat Terrain, notably, HyperR-MoMo obtains lower training performance, even though its average test performance is between ModuMorph and R-MoMo. In Incline, HyperR-MoMo shows the lowest training performance, and a similar test performance as ModuMorph. The lower performance and large variance across runs is due to a few runs in which the agent did not learn at all. 
This could be caused by larger instability due to the fact that the base network can contain a different set of parameters at every timestep (instead of the same set of parameters for each individual robot, as previously discussed).
These results indicate that this architecture variation might require additional hyperparameter optimization to reduce instability, but could be promising for universal morphology control.
Specifically because its generalization performance is still better than ModuMorph.

\subsection{Gated Recurrent Unit}
\begin{figure}[h]
    \centering
    \includegraphics[width=0.6\textwidth]{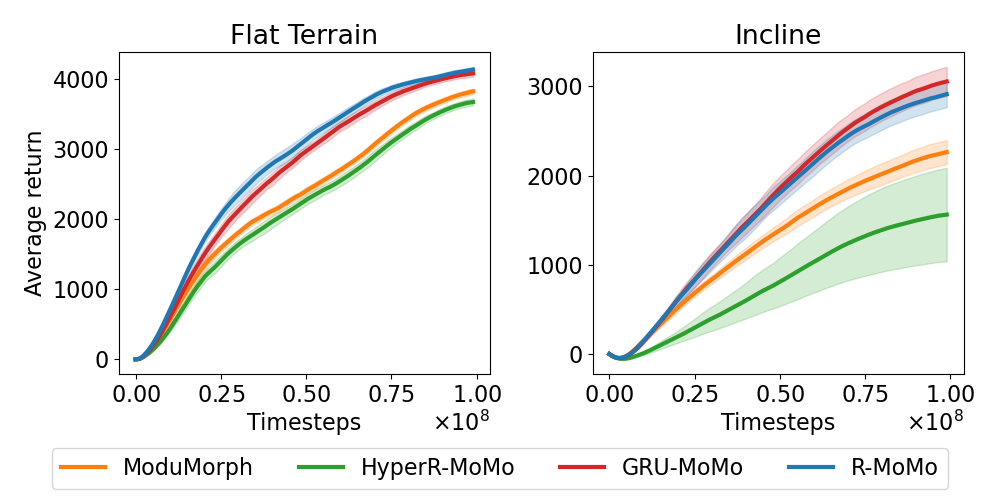}
    \caption{Training performance on the 100 training robots in Flat Terrain and Incline for ModuMorph, HyperR-MoMo, GRU-MoMo (R-MoMo with a GRU), and R-MoMo (with an LSTM). Average return with a 95\% confidence interval over 10 seeds is shown.}
    \label{fig:ablation_training_performance}
\end{figure}

\begin{figure}[h]
    \centering
    \includegraphics[width=0.6\textwidth]{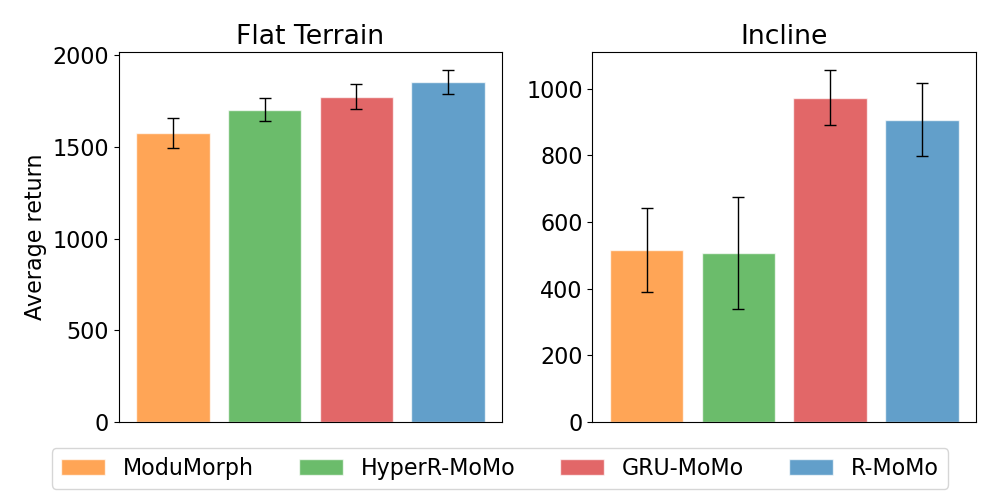}
    \caption{Test performance on the 70 unseen test robots in Flat Terrain and Incline for ModuMorph, HyperR-MoMo, GRU-MoMo (R-MoMo with a GRU), and R-MoMo (with an LSTM). Average return with a 95\% confidence interval over 10 seeds is shown.}
    \label{fig:ablation_test_performance}
\end{figure}

To assess the robustness of the performance gains with shared modular recurrence and to evaluate whether they are specific for LSTMs, we experimented with a GRU. Leaving the rest of the architecture unchanged, the LSTM is swapped with a GRU in R-MoMo. The training performance in Figure \ref{fig:ablation_training_performance} and test performance in Figure \ref{fig:ablation_test_performance} in Flat Terrain and Incline show that this \textit{GRU-MoMo} is competitive with R-MoMo that uses an LSTM. These results demonstrate that the performance gains are not dependent on the usage of LSTM and provide additional evidence of the necessity of inferring latent contextual features from the history.

\section{Comparisons}\label{appendix:comparisons}

\subsection{Reward Per Timestep}

\begin{figure}[htbp]
    \centering
    \includegraphics[width=\linewidth]{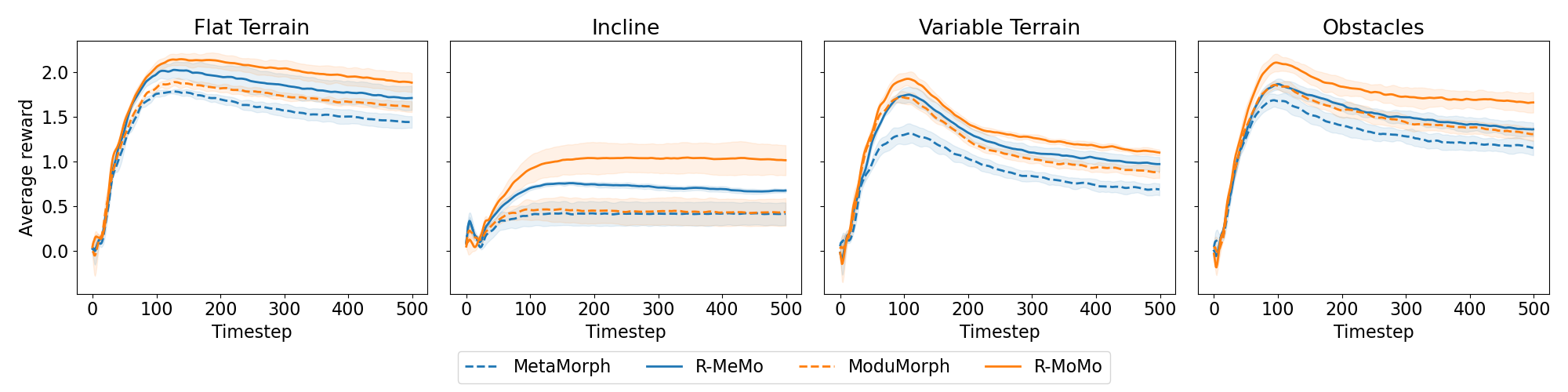}
    \caption{The reward obtained at every timestep in the first half of the episode in the different environments. The obtained rewards are averaged over all test robots and 5 seeds and the standard deviation is shown.}
    \label{fig:reward_per_t}
\end{figure}

To qualitatively ensure that the observed performance gains are not the result of a limited number of transitions with a large reward (for example at the very beginning of the episode), we analyzed the reward obtained by the agent at every timestep during an episode. In Figure \ref{fig:reward_per_t}, the obtained average reward per timestep for all test robots is shown for the first 500 timesteps of the episode. In general, the reward slowly increases at the start of the episode when the agent has to apply torques to induce locomotion. In Variable Terrain and Obstacles, the average reward per timestep decreases as soon as the terrain changes or obstacles arise, as these slow down the robot. The differences between the methods demonstrate that the performance improvement with shared modular recurrence over the baselines is consistent over timesteps during the episode.

\begin{figure}[htbp]
    \centering
    \includegraphics[width=\linewidth]{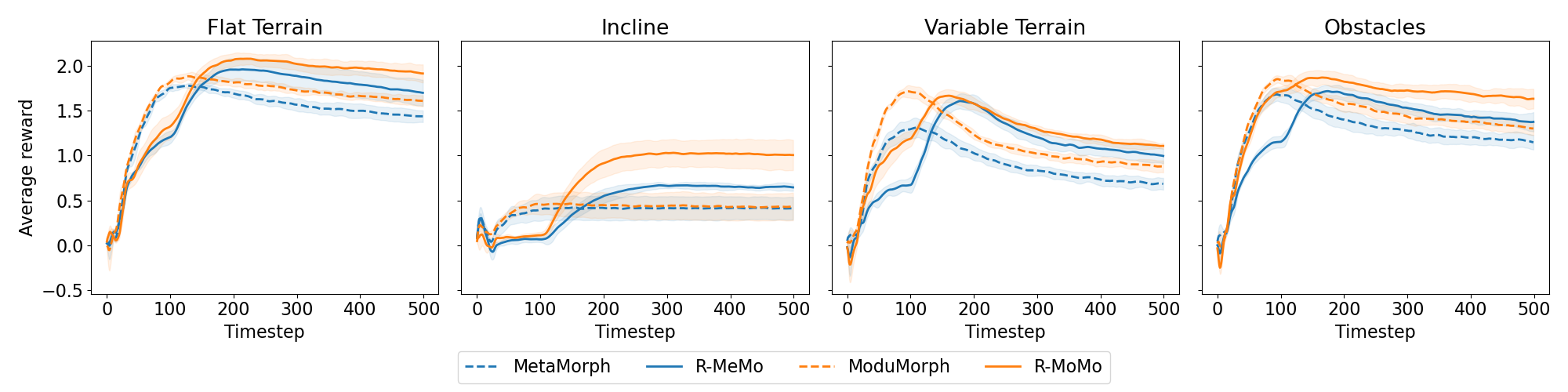}
    \caption{The reward obtained at every timestep in the first half of the episode in the different environments, where the recurrent agents receive zeros as hidden states during the first 100 timesteps. The obtained rewards are averaged over all test robots and 5 seeds and the standard deviation is shown.}
    \label{fig:reward_per_t_zeros}
\end{figure}

Additionally, to investigate the agent's usage of the RNN hidden states, we performed the same analysis for recurrent agents that receive hidden states consisting of only zeros during the first 100 timesteps of each episode. In Figure \ref{fig:reward_per_t_zeros} we can now see that the recurrent agents execute a weak policy during these timesteps, but are able to recover up to the original reward per timestep observed in Figure \ref{fig:reward_per_t}. This demonstrates that the hidden states are encoding crucial information.

\subsection{Wall-Clock Training Time}

In Table \ref{tab:wallclocktime} the wall-clock time of the different methods for 100M environmental steps and training in Flat Terrain is reported (note that this includes logging to different destinations and storing model checkpoints). Experiments were performed on NVIDIA A40 GPUs of the \citet{DAIC}. 
Chunking the collected episodes before training, and applying the burn-in strategy, introduced by \citet{r2d2} for DRQN, reduces the extra overhead of sequential processing by the RNNs. Still, training time is increased significantly. However, this difference disappears during inference as data is then processed sequentially for all methods.
The trade-off between performance and the reduction of the chunk and/or burn-in size could be further exploited to reduce training time, which we leave for future work.

\begin{table}[htbp]
    \centering
    \caption{The wall-clock time of the different methods for 100M timesteps of training on Flat Terrain, averaged over 10 seeds.}
    \begin{tabular}{lc}
        
        \toprule
         & \textbf{Wall-clock training time} \\
        
        \midrule
        MetaMorph & $16.3 \pm 0.5$ hours \\
        R-MeMo & $21.8 \pm 0.9$ hours \\
        ModuMorph  & $21.9 \pm 0.6$ hours \\
        R-MoMo & $26.0 \pm 1.1$ hours \\
        \bottomrule
        
    \end{tabular}
    
    \label{tab:wallclocktime}
\end{table}

\begin{figure}[htbp]
    \centering
    \includegraphics[width=0.6\linewidth]{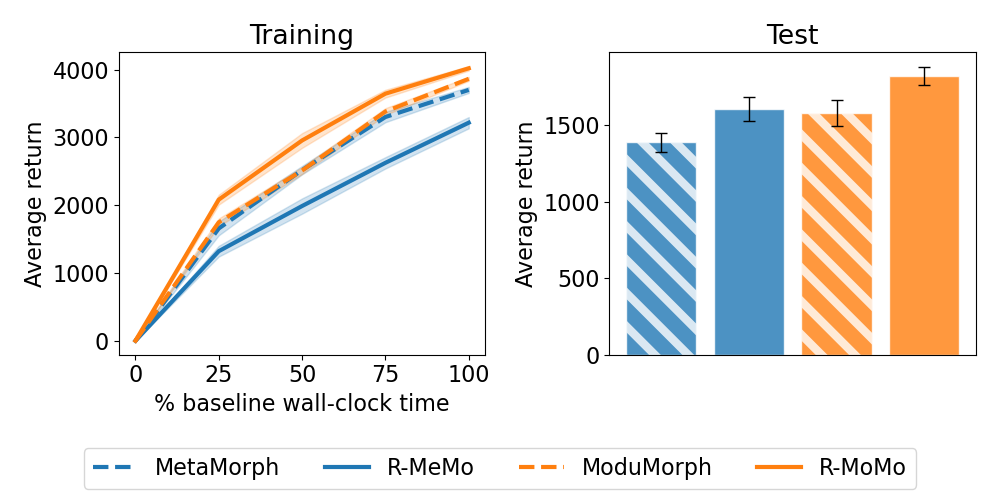}
    \caption{Training (left) and final test (right) performance in Flat Terrain when the recurrent methods receive the same wall-clock training time budget as their baselines (16.3 hours for R-MeMo and 21.9 hours for R-MoMo). Results are averaged over 10 seeds and 95\% CIs are shown.}
    \label{fig:wallclock_curves}
\end{figure}

As an additional comparison between the baselines, MetaMorph and ModuMorph, and their recurrent counterparts, R-MeMo and R-MoMo, we report results where each recurrent method is given the same training time (wall-clock time) as its baseline. Thus, R-MeMo is trained for 16.3 hours and R-MoMo is trained for 21.9 hours.
Figure \ref{fig:wallclock_curves} shows training performance over these wall-clock times on the left, and final test performance, as measured with the unseen robot topologies, on the right, in the Flat Terrain environment.
Comparing the training results with the original results in Figure \ref{fig:train_performance}, it can be observed that MetaMorph now reaches higher training performance than R-MeMo, and the difference between ModuMorph and R-MoMo is smaller.
However, even though the training times of R-MeMo and R-MoMo are cut short, the differences in final test performance remain similar to those observed under the full budget of 100M timesteps (in Figure \ref{fig:test_returns}).

\subsection{Zero-Shot Generalization Performance Improvement Per Environment}

\begin{figure}[htbp]
    \centering    
    \begin{subfigure}[b]{0.7\textwidth}
        \includegraphics[width=\textwidth]{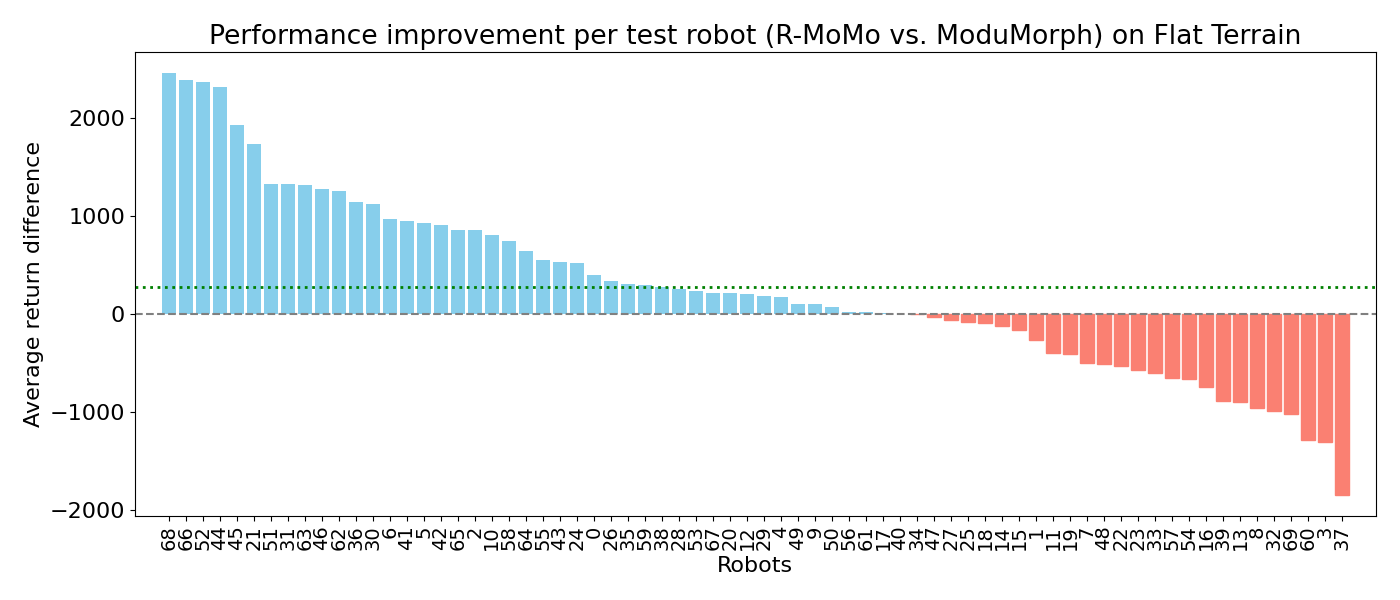}
    \end{subfigure}
    \begin{subfigure}[b]{0.7\textwidth}
        \includegraphics[width=\textwidth]{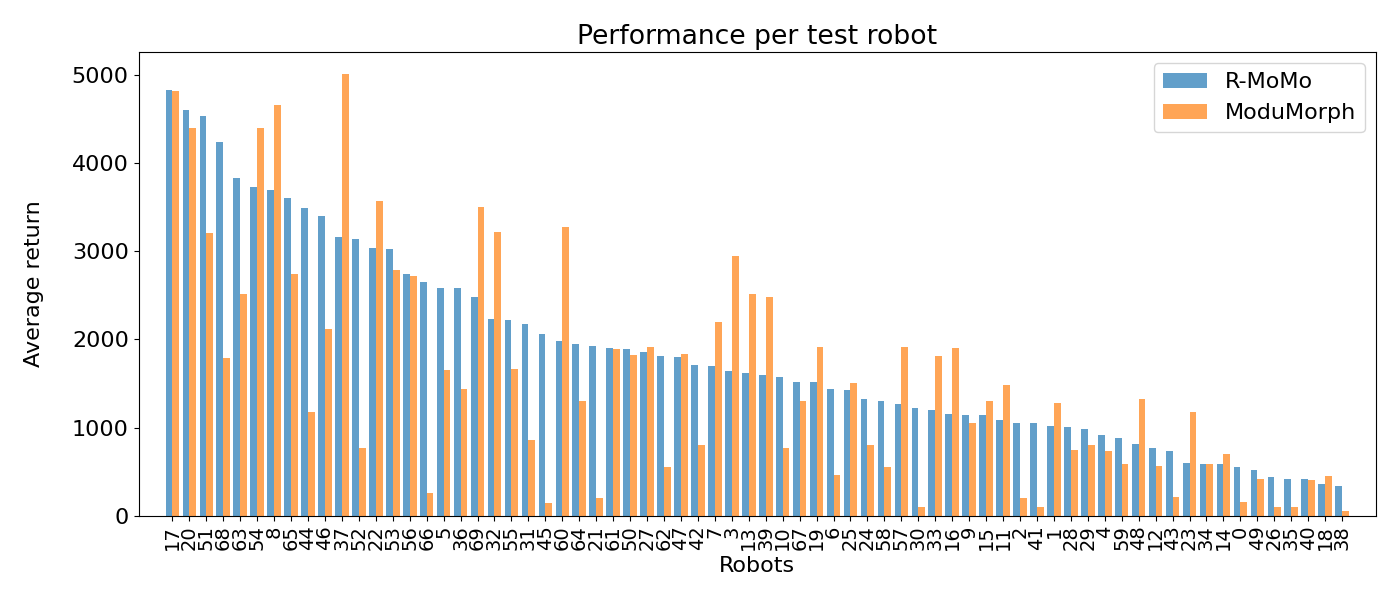}        
    \end{subfigure}
    \caption{
    The difference in return between R-MoMo and ModuMorph (top) and the obtained returns (bottom) on each of the 70 unseen test robots in the \textbf{Flat Terrain} environment. Returns are averaged over 10 seeds.}
    \label{fig:per_robot_test_ft}
\end{figure}

\begin{figure}[htbp]
    \centering    
    \begin{subfigure}[b]{0.7\textwidth}
        \includegraphics[width=\textwidth]{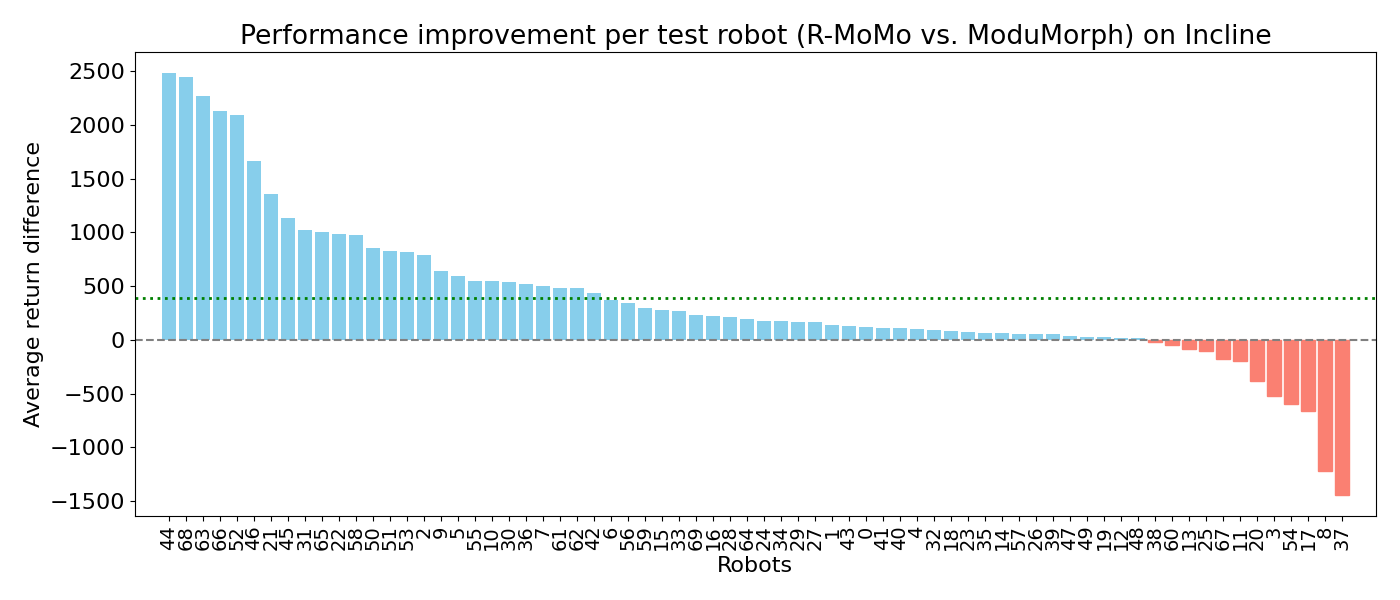}
    \end{subfigure}
    \begin{subfigure}[b]{0.7\textwidth}
        \includegraphics[width=\textwidth]{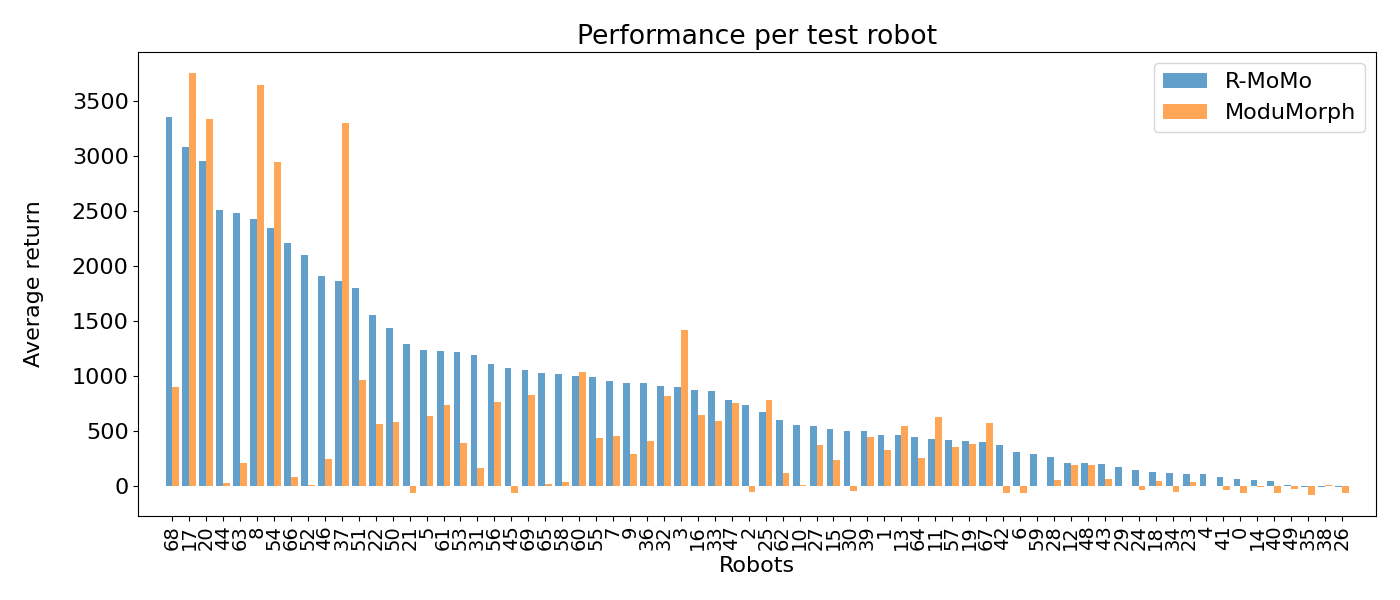}        
    \end{subfigure}
    \caption{
    The difference in return between R-MoMo and ModuMorph (top) and the obtained returns (bottom) on each of the 70 unseen test robots in the \textbf{Incline} environment. Returns are averaged over 10 seeds.}
    \label{fig:per_robot_test_incline}
\end{figure}

\begin{figure}[htbp]
    \centering    
    \begin{subfigure}[b]{0.7\textwidth}
        \includegraphics[width=\textwidth]{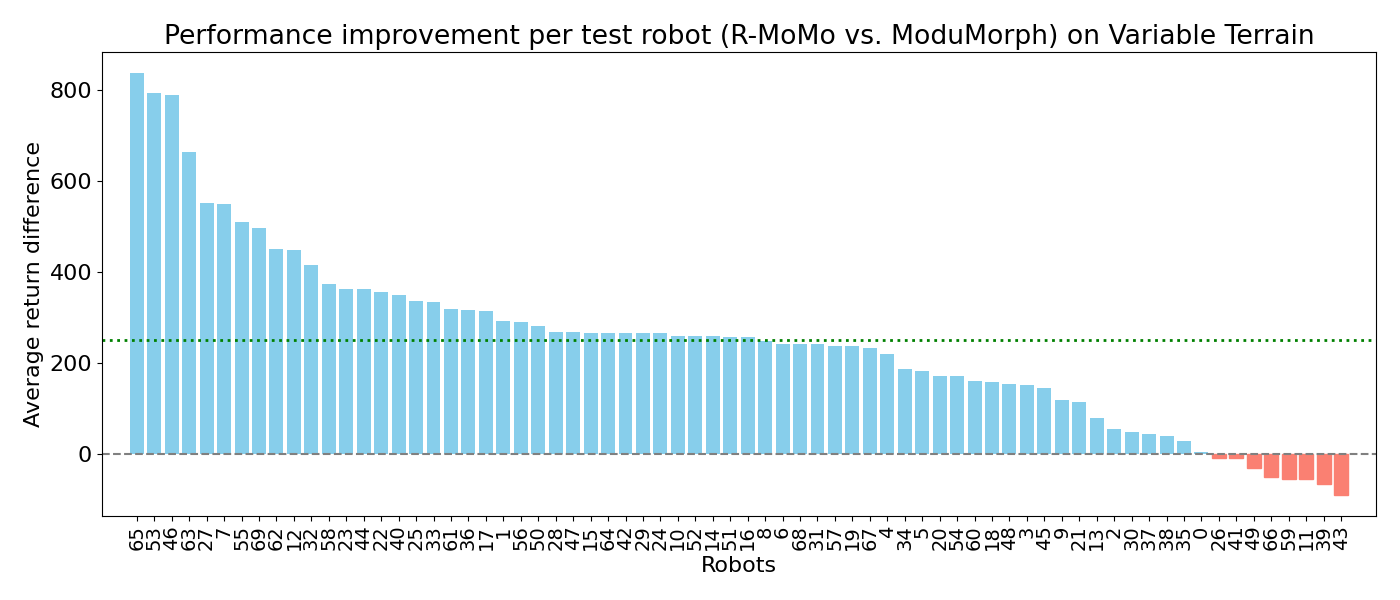}
    \end{subfigure}
    \begin{subfigure}[b]{0.7\textwidth}
        \includegraphics[width=\textwidth]{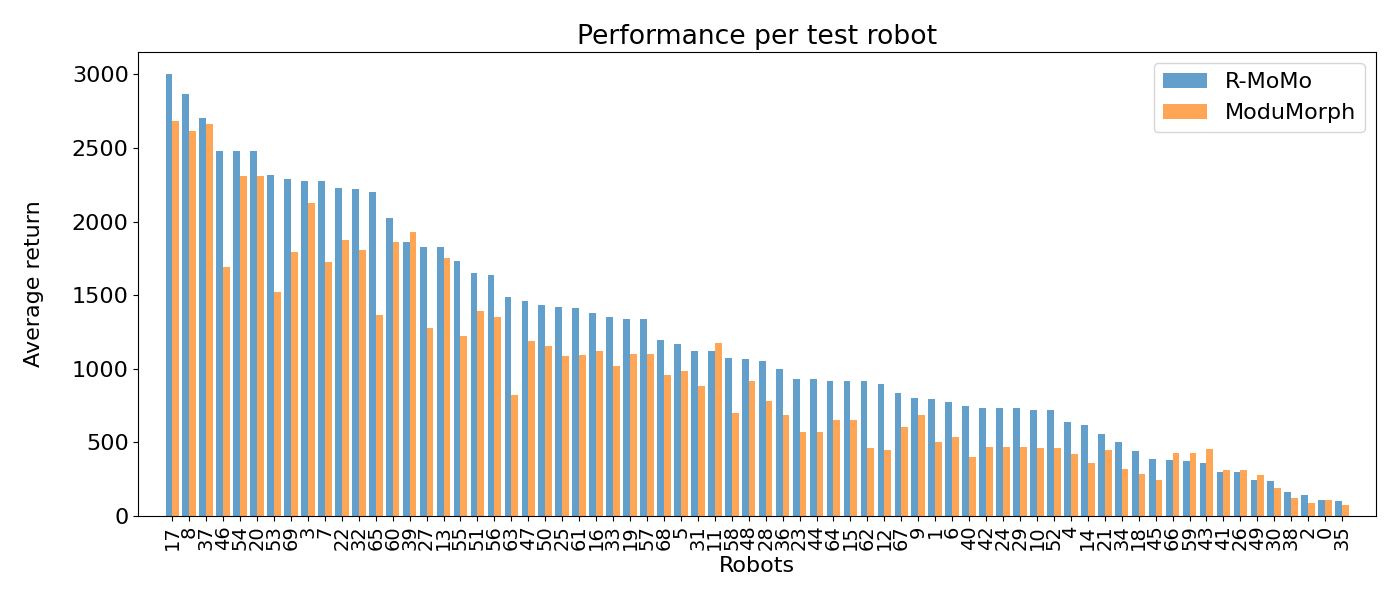}        
    \end{subfigure}
    \caption{
    The difference in return between R-MoMo and ModuMorph (top) and the obtained returns (bottom) on each of the 70 unseen test robots in the \textbf{Variable Terrain} environment. Returns are averaged over 10 seeds.}
    \label{fig:per_robot_test_csr}
\end{figure}

\begin{figure}[htbp]
    \centering    
    \begin{subfigure}[b]{0.7\textwidth}
        \includegraphics[width=\textwidth]{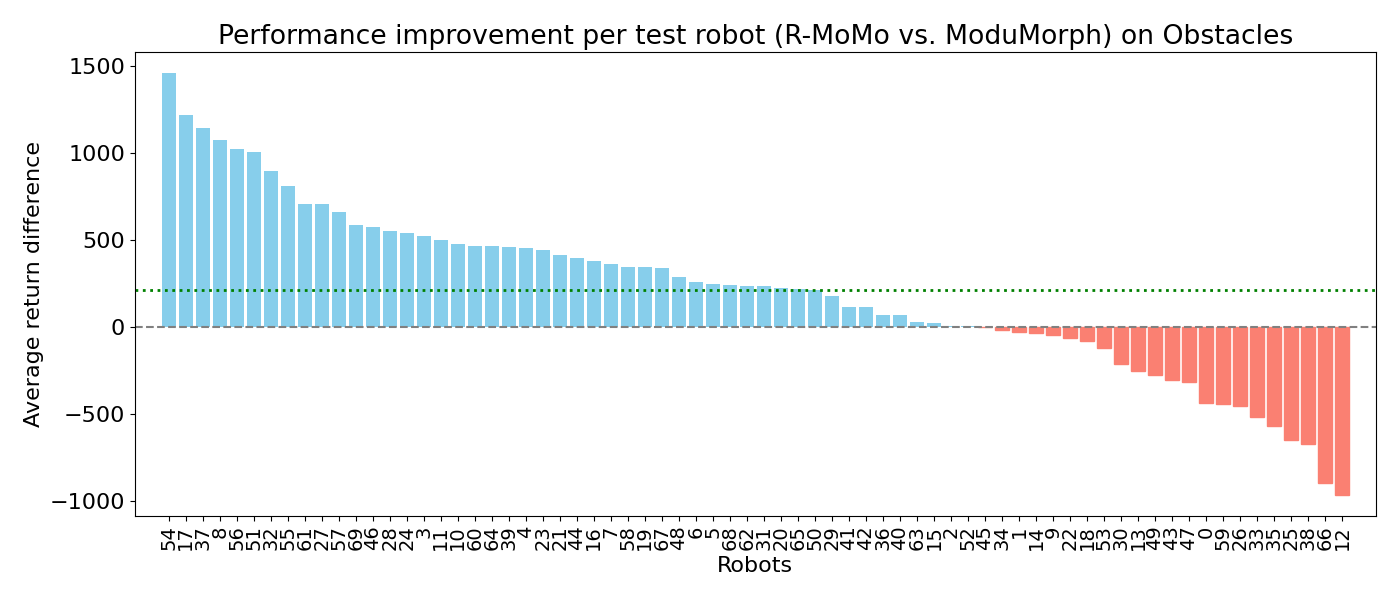}
    \end{subfigure}
    \begin{subfigure}[b]{0.7\textwidth}
        \includegraphics[width=\textwidth]{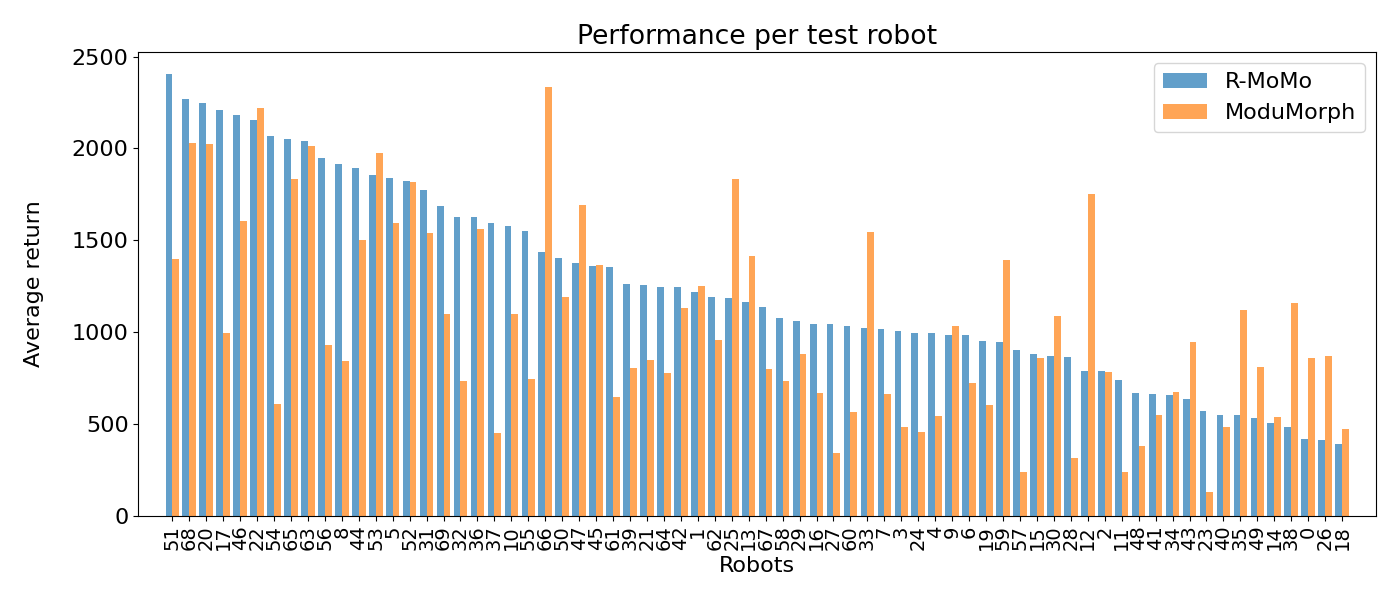}        
    \end{subfigure}
    \caption{
    The difference in return between R-MoMo and ModuMorph (top) and the obtained returns (bottom) on each of the 70 unseen test robots in the \textbf{Obstacles} environment. Returns are averaged over 10 seeds.}
    \label{fig:per_robot_test_obstacles}
\end{figure}

The difference in performance on test robots between R-MoMo and ModuMorph is shown in Figures \ref{fig:per_robot_test_ft}, \ref{fig:per_robot_test_incline}, \ref{fig:per_robot_test_csr} and \ref{fig:per_robot_test_obstacles} for the Flat Terrain, Incline, Variable Terrain and Obstacles environments, respectively. These results show increased test performance for R-MoMo on a majority of the test robots across all environments. Moreover, R-MoMo generally struggles with fewer robots (e.g. return below 500) than ModuMorph.

\end{document}